\documentclass[11pt]{article}

\usepackage[final]{acl}

\usepackage{times}
\usepackage{dsfont}
\usepackage{latexsym}
\usepackage{microtype}
\usepackage{amssymb}
\usepackage{tcolorbox}
\usepackage{amsmath}
\usepackage{fdsymbol}
\usepackage{bbm}
\usepackage{inconsolata}
\usepackage{float}
\usepackage{multirow}
\usepackage{graphicx}
\usepackage{tikz}
\usepackage{booktabs}
\usepackage{caption}
\usepackage{subcaption}
\usepackage{xspace}
\usepackage{makecell}
\usepackage{enumerate}
\usepackage{booktabs}
\usepackage[capitalize,noabbrev]{cleveref}
\usepackage{tcolorbox}
\usepackage[T1]{fontenc}
\usepackage{listings}
\usepackage{xcolor}
\usepackage{colortbl}
\usepackage{pifont}

\usepackage[utf8]{inputenc}

\newcommand\eg{\emph{e.g.},\xspace}

\newcommand{\papertitle}{\textbf{LegalSearch-R1}\xspace}

\definecolor{stepcolor}{HTML}{d79b00}
\definecolor{contentcolor}{HTML}{6c8ebf}

\definecolor{cYellow}{RGB}{255,255,3}
\definecolor{cBlue}{RGB}{69,123,157}
\definecolor{cRed}{RGB}{231,56,71}
\definecolor{cRed_1}{RGB}{191,30,46}
\definecolor{cGray}{RGB}{168,218,219}
\definecolor{cBlue_2}{RGB}{5,48,97}
\definecolor{cBlue_1}{RGB}{115,186,214}
\definecolor{cBlue_3}{RGB}{13,76,109}
\definecolor{cBlue_4}{RGB}{64,121,160}
\definecolor{cOrange}{RGB}{250,134,0}
\definecolor{cBlue_6}{RGB}{13,76,109}
\definecolor{cBlue_7}{RGB}{16,106,130}
\definecolor{cBlue_8}{RGB}{19,136,160}
\definecolor{cBlue_9}{RGB}{115,184,214}
\title{Can LLMs Time Travel? Enhancing Temporal Consistency in Legal Agentic Search through Reinforcement Learning}

\author{
Wei Fan\textsuperscript{1},
Yining Zhou\textsuperscript{1},
Mufan Zhang\textsuperscript{1},
Yanbing Weng\textsuperscript{1},
Yiran Hu\textsuperscript{3},
Tianshi Zheng\textsuperscript{1},\\
\textbf{
Baixuan Xu\textsuperscript{1},
Chunyang Li\textsuperscript{1},
Jianhui Yang\textsuperscript{2},
Haoran Li\textsuperscript{1}\thanks{Corresponding Authors},
Yangqiu Song\textsuperscript{1}}\\
\textsuperscript{1}Department of Computer Science and Engineering, HKUST, Hong Kong SAR, China\\
\textsuperscript{2}School of Law, Tsinghua University, Beijing, China\\
\textsuperscript{3}Cheriton School of Computer Science, University of Waterloo, Waterloo, Canada\\
\texttt{wfanag@connect.ust.hk}, \texttt{hlibt@connect.ust.hk}, \texttt{yqsong@cse.ust.hk}\\
}

\begin{document}
\maketitle

\begin{abstract}
While large language models~(LLMs) augmented with agentic search capabilities show promise for legal reasoning, they overlook a fundamental constraint that applicable law must match the temporal context of each case, as retroactive application of statutes violates core legal principles and leads to erroneous conclusions.
Our observations reveal that current legal LLMs suffer from temporal bias anchored to their training cutoff, while search agents rarely incorporate temporal constraints into queries, and that web search alone cannot provide the precise statute and precedent citations that legal reasoning demands.
To address these challenges, we propose \papertitle, an end-to-end reinforcement learning framework that pairs local statute RAG for precise article matching with online web search for broader legal knowledge, trained on temporally-indexed data spanning multiple amendment periods to enforce temporal consistency.
Extensive experiments on our benchmark covering 13 legal tasks demonstrate that our 7B-parameter agent outperforms state-of-the-art deep research frameworks and specialized legal LLMs by 12.9\% to 29.8\%, surpasses baselines by 57.7\% to 80.3\% on temporal consistency, and exhibits robust out-of-domain generalization.
The code and data are available at \url{https://github.com/HKUST-KnowComp/LegalSearch-R1}.
\end{abstract}

\begin{quote}
\centering
\emph{``A retroactive law is truly a monstrosity.'' \\[0.5ex]
(Lex retro non agit)} \\[1ex]
{\raggedleft\small--- Lon L. Fuller, \textit{The Morality of Law} (1969)\par}
\end{quote}

\begin{figure}[t!]
    \centering
    \includegraphics[width=0.47\textwidth]{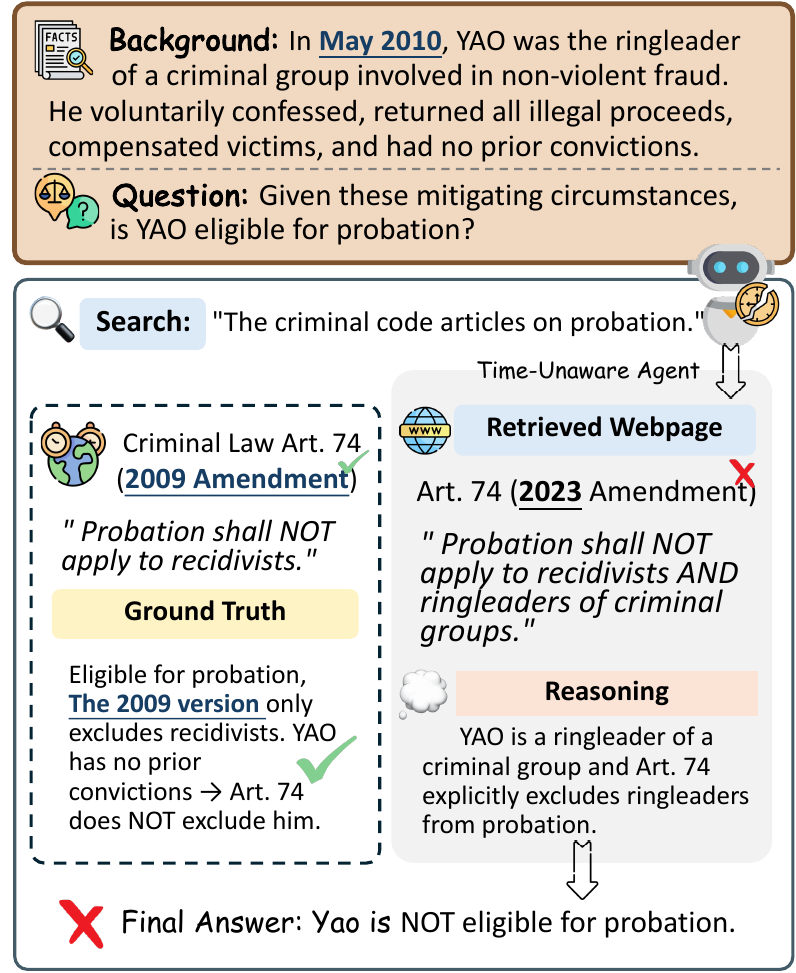}
    \caption{Temporal inconsistency in legal reasoning, where the same case facts yield opposite probation conclusions under the 2009 vs. 2023 amendments of Article~74 of the Chinese Criminal Law.}
    \vspace{-0.2in}
    \label{figs:introduction}
\end{figure}

\section{Introduction}

Legal reasoning operates through the application of normative rules derived from statutes and judicial precedents~\cite{guha2023legalbenchcollaborativelybuiltbenchmark,fei2023lawbenchbenchmarkinglegalknowledge}. Whether grounded in codified provisions of civil law systems or the binding precedents of common law traditions, practitioners must identify the legal authority applicable to each case and ensure its temporal alignment with the facts at issue.

Large language models~(LLMs)~\cite{openai2024gpt4technicalreport,qwen2025qwen3technicalreport,deepseekai2025deepseekv3technicalreport} and legal-specialized variants~\cite{cui2023chatlaw,yue2023disclawllm,wu2023fuzimingcha,dai2025legaldelta} advance legal reasoning tasks such as judgment prediction~\cite{zhong2020iteratively}, question answering~\cite{louis2024interpretable}, and statute retrieval~\cite{fei2023lawbenchbenchmarkinglegalknowledge}.
To overcome the knowledge boundary of static pre-training~\cite{fan-etal-2024-goldcoin}, RAG~\cite{lewis2021retrievalaugmentedgenerationknowledgeintensivenlp,gao2024retrievalaugmentedgenerationlargelanguage} and deep research agents~\cite{openaideepresearch,geminideepresearch,zheng2025deepresearcherscalingdeepresearch} combine multi-step reasoning~\cite{deepseekai2025deepseekr1incentivizingreasoningcapability,openai2026openaio1card} with action generation~\cite{yao2023reactsynergizingreasoningacting}, and RL-based search agents~\cite{jin2025searchr1trainingllmsreason,song2025r1searcherincentivizingsearchcapability,zhang2025evolvesearchiterativeselfevolvingsearch} demonstrate strong results on open-domain benchmarks.

However, applying these systems to law exposes two critical challenges that existing work largely overlooks.
The first is \textit{temporal consistency}.
Legal knowledge is subject to discrete, authoritative amendments, and applicable law must match the temporal context of each case.
As illustrated in~\cref{figs:introduction}, a 2010 case must be evaluated under the 2009 amendment of Article~74 of the Chinese Criminal Law; applying the 2023 amendment leads to the opposite conclusion.
The principle of \textit{lex retro non agit} holds that applicable law should align with the temporal context of each case, as ``a retroactive law is truly a monstrosity''~\cite{fuller1969morality,wang2024lekube}.
Our observations reveal that current LLMs suffer from temporal bias anchored to their training cutoff~\cite{t-y-s-s-etal-2024-chronoslex,t-y-s-s-vuong-2025-lextempus}. As shown in~\cref{figs:heatmap}, all evaluated models peak on provisions near their training cutoff (2021--2022) while degrading on both earlier and post-cutoff versions. Meanwhile, search agents rarely incorporate temporal constraints when formulating queries as shown in~\cref{figs:temporal_query}.
The second is \textit{precise legal knowledge retrieval}.
Legal reasoning requires authoritative domain knowledge, including statutory provisions, judicial precedents, and established legal principles, with article-level precision. Web search excels at locating authoritative analyses but is inadequate for retrieving precise statutory text, as results are noisy, frequently truncating provisions and mixing them with commentary. While legal RAG systems have been explored for case retrieval~\cite{li2023sailerstructureawarepretrainedlanguage,ma2021lecard} and legal document generation~\cite{park2025lragelegalretrievalaugmented}, they operate as static pipelines without temporal filtering, making them unable to distinguish between superseded and current provisions.

To address these, we propose \papertitle, an end-to-end reinforcement learning framework with two complementary mechanisms.
First, we construct temporally-indexed training data spanning multiple amendment periods and train the agent via GRPO~\cite{shao2024deepseekmathpushinglimitsmathematical} to identify the temporal context of each query, directly overcoming parametric temporal bias.
Second, we equip the agent with a dual-tool architecture pairing local RAG over a curated statute corpus with online web search for broader judicial interpretations.
We further adopt entropy-based advantage shaping~\cite{fan2025deepplannerscalingplanningcapability} to accelerate the learning of temporal query formulation, a planning-stage decision critical to legal search.

Extensive experiments on our benchmark covering 7 in-domain and 6 out-of-domain legal tasks demonstrate that our 7B agent outperforms state-of-the-art open-source deep research agents and specialized legal LLMs by 12.9\% to 29.8\%, surpasses all baselines by 57.7\% to 80.3\% on temporal consistency, and exhibits robust out-of-domain generalization.
Our contributions are as follows:
\begin{itemize}
\item We identify and empirically verify \textit{temporal inconsistency} as a critical yet overlooked failure mode in legal agentic search, where both legal LLMs and search agents systematically neglect the temporal context of legal queries.
\item We propose \papertitle, the first end-to-end RL framework synergizing local statute RAG with online web search, trained on temporally-indexed data with advantage shaping that accelerates temporal query planning.
\item We construct a temporally-indexed benchmark covering 13 legal tasks and demonstrate that \papertitle significantly outperforms larger models and existing deep research frameworks on both in-domain and out-of-domain evaluations.
\end{itemize}

\section{Related Work}
\subsection{Large Language Models in Law}
Legal-specialized LLMs emerge through domain-adaptive pre-training and instruction tuning, including ChatLaw~\cite{cui2023chatlaw}, Lawyer LLaMA~\cite{huang2023lawyerllama}, DISC-LawLLM~\cite{yue2023disclawllm}, LawGPT~\cite{zhou2024lawgpt}, InternLM-Law~\cite{fei-etal-2025-internlm}, and SaulLM~\cite{colombo2024saullm} for English. Recent work explores RL for legal reasoning, including LegalDelta~\cite{dai2025legaldelta}, Unilaw-R1~\cite{cai2025unilawR1}, and SyLeR~\cite{zhang2025syler}.
Despite these advances, all legal LLMs share a fundamental limitation: parametric knowledge anchored to the training cutoff becomes temporally misaligned as statutes undergo periodic amendments~\cite{wang2024lekube}. ChronosLex~\cite{t-y-s-s-etal-2024-chronoslex} and LexTempus~\cite{t-y-s-s-vuong-2025-lextempus} reveal significant degradation on temporally shifted legal inputs, and FreshLLMs~\cite{vu2024freshllms} and LexTime~\cite{barale2025lextime} confirm that LLMs broadly struggle with time-sensitive knowledge.
Retrieval-augmented approaches such as SAILER~\cite{li2023sailerstructureawarepretrainedlanguage} for legal case retrieval, LeCaRD~\cite{ma2021lecard} for benchmarking Chinese legal retrieval, and recent RAG evaluation tools~\cite{park2025lragelegalretrievalaugmented} have been explored, but these systems treat retrieval as a static, one-shot process without temporal filtering or multi-turn reasoning.
These gaps motivate integrating agentic search to dynamically retrieve temporally appropriate provisions with precise, article-level statute matching.

\subsection{Agentic Search for Legal Reasoning}
Agentic search systems ground LLM outputs in up-to-date sources through external tool use. Open Deep Search~\cite{alzubi2025opendeepsearch} enhances search through carefully designed interaction patterns, but the hand-engineered workflows limit adaptability.
RL-based approaches offer end-to-end training: Search-R1~\cite{jin2025searchr1trainingllmsreason}, R1-Searcher~\cite{song2025r1searcherincentivizingsearchcapability}, and DeepResearcher~\cite{zheng2025deepresearcherscalingdeepresearch} train LLMs via outcome-based RL to interleave multi-turn search with reasoning. EvolveSearch~\cite{zhang2025evolvesearchiterativeselfevolvingsearch} iterates between SFT and RL, ReSearch~\cite{chen2025researchlearningreasonsearch} treats search as integral to reasoning, WebThinker~\cite{li2025webthinker} enables autonomous web navigation, and DeepPlanner~\cite{fan2025deepplannerscalingplanningcapability} introduces advantage shaping to scale planning capacity. In the legal domain, LRAS~\cite{zhou2026lras} and L-MARS~\cite{wang2026lmarslegalmultiagentworkflow} demonstrate significant gains by transitioning legal LLMs from closed-loop reasoning to dynamic agentic search.
While these agents achieve impressive results, none enforces temporal consistency across legal queries during training, and they rely on web search without structured statute retrieval for precise article matching, a gap our framework fills by pairing temporally-indexed training data with a hybrid architecture of local statute RAG and online web search.

\begin{figure}[t!]
    \centering
    \includegraphics[width=0.47\textwidth]{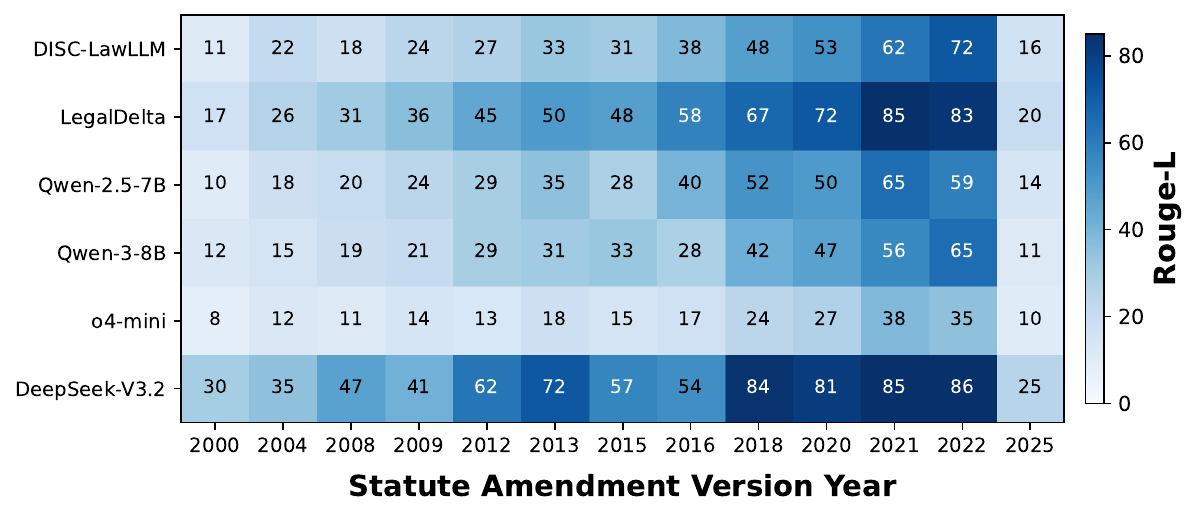}
    \caption{LAR ROUGE-L scores across amendment version years. All models peak near their training cutoff (2021--2022) and degrade on post-cutoff provisions.}
    \vspace{-0.1in}
    \label{figs:heatmap}
\end{figure}

\begin{figure}[t!]
    \centering
    \includegraphics[width=0.47\textwidth]{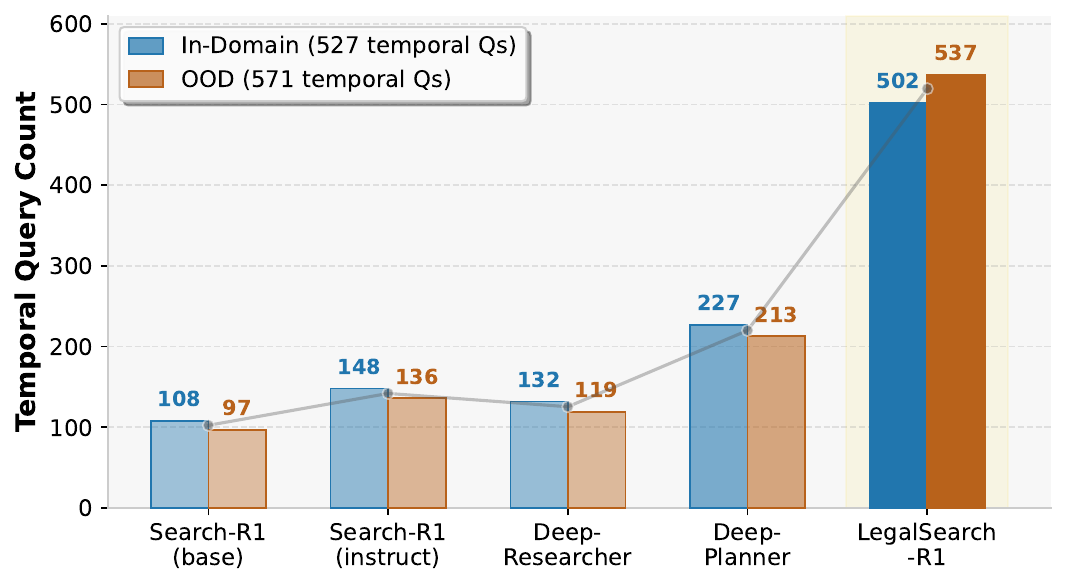}
    \caption{Number of temporally-aware search queries. Baselines rarely include temporal constraints, while \papertitle consistently does.}
    \vspace{-0.2in}
    \label{figs:temporal_query}
\end{figure}

\section{Preliminaries}

\subsection{Task Formulation}
In the legal agentic search scenario, the agent receives a temporally-contextualized legal query $(q \mid t_q)$, where $q$ is the legal question and $t_q \in \mathcal{Y}$ denotes the temporal context of the query, typically the year in which the legal events occurred. The temporal context $t_q$ is critical because it determines which version of the applicable statute governs the case, and the same legal question may yield entirely different answers under different statutory versions. For instance, a query about probation eligibility under Article~74 of the Chinese Criminal Law produces opposing conclusions when evaluated under $t_q = 2009$ versus $t_q = 2023$ (\cref{figs:introduction}). Given $(q \mid t_q)$ and a system prompt $p$ (detailed in~\cref{app:system_prompt}), the LLM-based agent follows the ReAct~\cite{yao2023reactsynergizingreasoningacting} framework to perform multi-turn legal reasoning and action until producing a final answer grounded in authoritative legal sources. Following~\citet{fan2025deepplannerscalingplanningcapability}, we formalize the interaction trajectory as:
\begin{equation} \label{eq:trajectory}
\small
\tau = \{(s_0, e_0, a_0), (s_1, e_1, a_1), \dots, (s_{T}, e_T, a_{T}), R\},
\end{equation}
where the state $s_t$ at step $t$ is defined as:
\begin{equation} \label{eq:state}
\small
s_t = (p,\; q,\; t_q,\; \{e_i\}_{i=0}^{t},\; \{a_i\}_{i=0}^{t},\; \{o_i\}_{i=0}^{t-1}),
\end{equation}
comprising the system prompt $p$, legal query $q$, temporal context $t_q$, accumulated reasoning segments $\{e_i\}$, action tokens $\{a_i\}$, and tool responses $\{o_i\}$.
At each step, the model generates a thinking segment $e_t$ and an action $a_t$ that corresponds to either a high-level plan or a low-level execution. $R$ denotes the terminal reward that evaluates the quality of the final answer with respect to the legal query $(q \mid t_q)$.

\begin{figure*}[t]
    \centering
    \includegraphics[width=2\columnwidth]{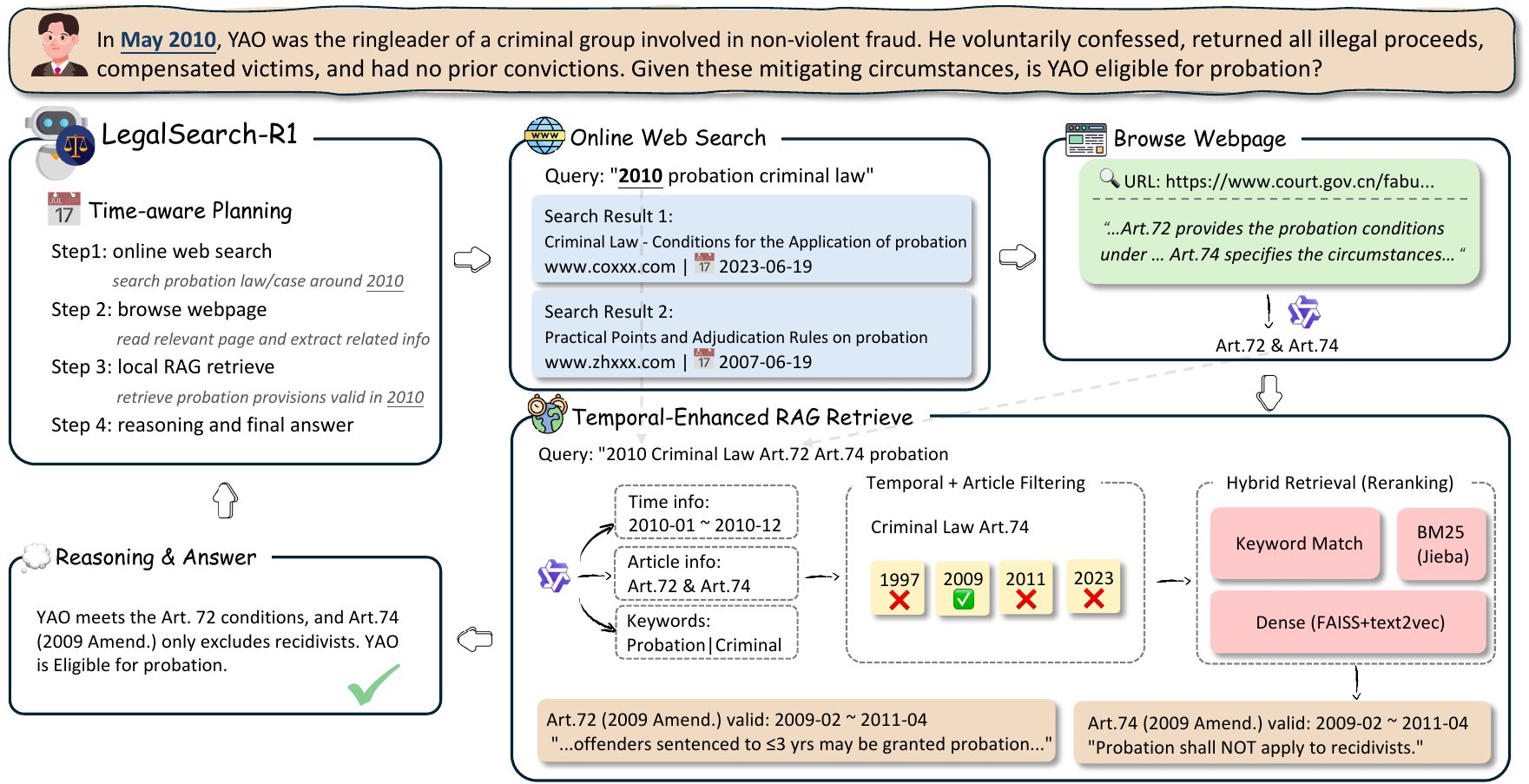}
    \caption{Overview of \papertitle. The agent performs multi-turn legal reasoning over a hybrid online-local retrieval architecture. The temporal-enhanced RAG module first analyzes the query via an LLM to extract temporal references and legal keywords, then applies temporal filtering to restrict candidates to provisions valid at the query time, and finally aggregates keyword, dense vector, and BM25 retrieval channels via Reciprocal Rank Fusion.}
    \label{figs:method}
    \vspace{-0.2in}
\end{figure*}

\subsection{Basic Agentic Modules}
At each step, the agent output decomposes into four modules. \textbf{Think} generates a reasoning segment wrapped in \texttt{<think>}$\cdots$\texttt{</think>} tags, serving as the agent's legal analysis of the current state $s_t$. \textbf{Plan} produces a high-level legal research strategy enclosed in \texttt{<plan>}$\cdots$\texttt{</plan>} tags that specifies what statutory provisions or legal knowledge to retrieve, which tools to invoke, and how to integrate the evidence; an initial plan is required in the first round and may be refined in subsequent steps. \textbf{Tool Call} invokes one of the available tools $\mathcal{T} = \{\texttt{web\_search},\; \texttt{rag\_retrieve},\; \texttt{browse\_webpage}\}$ via a structured request in \texttt{<tool\_call>} tags, following the tool schema specified in~\cref{app:tool_schema}. Together, \texttt{web\_search} and \texttt{rag\_retrieve} form a hybrid online-local retrieval architecture, where online web search provides broad access to judicial interpretations and legal commentary while local statute RAG ensures precise, version-aware article retrieval. The environment returns a tool response $o_t$ appended to the agent context for the next step. \textbf{Answer} is the termination action wrapped in \texttt{<answer>}$\cdots$\texttt{</answer>} tags, which triggers the terminal reward $R$.

\section{Temporally-Indexed Legal Tasks}\label{sec:legal_tasks}

\begin{table}[t]
\renewcommand{\arraystretch}{1.05}
\small
\setlength{\tabcolsep}{3pt}
\begin{center}
\begin{tabular}{l l r r r r}
\toprule
\multirow{2}{*}{\textbf{Task}} & \multirow{2}{*}{\textbf{Source}} & \multicolumn{2}{c}{\textbf{Train}} & \multicolumn{2}{c}{\textbf{Test}} \\
\cmidrule(lr){3-4} \cmidrule(lr){5-6}
& & \textbf{\#All} & \textbf{\#Temp.} & \textbf{\#All} & \textbf{\#Temp.} \\
\midrule
\multicolumn{6}{l}{\textit{In-Domain}} \\
KQA & LawBench & 512 & 11 & 128 & 5 \\
CCP & LawBench & 512 & 501 & 128 & 125 \\
PTP & LawBench & 512 & 501 & 128 & 125 \\
LCA & LawBench & 512 & 71 & 128 & 16 \\
LCS & LawBench & 512 & 15 & 128 & 5 \\
LAP & LexEval & 512 & 493 & 128 & 123 \\
LAR & Ours & 512 & 512 & 128 & 128 \\
\midrule
\multicolumn{6}{l}{\textit{Out-of-Domain}} \\
6 tasks & DISC-Law & -- & -- & 768 & 571 \\
\midrule
\textbf{Total} & & \textbf{3,584} & \textbf{2,104} & \textbf{1,664} & \textbf{1,098} \\
\bottomrule
\end{tabular}
\end{center}
\caption{Benchmark statistics. \textbf{\#Temp.} denotes the number of questions with explicit temporal context (mentioning a specific year). All in-domain tasks have both training and test splits, and OOD tasks are test-only.}
\label{tabs:benchmark_statistics}
\vspace{-0.1in}
\end{table}

We construct a temporally-indexed benchmark from over 100{,}000 candidate questions drawn from LawBench~\cite{fei2023lawbenchbenchmarkinglegalknowledge}, LexEval~\cite{li2024lexeval}, DISC-LawEval~\cite{yue2023disclawllm}, and our own annotations. The benchmark construction process explicitly reinforces temporal context in the selected questions, and the resulting statistics are summarized in~\cref{tabs:benchmark_statistics}. Representative examples for all 13 tasks are provided in~\cref{app:task_examples}.

\subsection{In-Domain Tasks.}
From LawBench, we select five tasks requiring legal reasoning grounded in external knowledge retrieval. \textbf{Knowledge Question Answering}~(KQA) tests factual legal knowledge via multiple-choice. \textbf{Criminal Charge Prediction}~(CCP) identifies applicable charges from case facts. \textbf{Prison Term Prediction}~(PTP) predicts sentencing duration. \textbf{Legal Case Analysis}~(LCA) selects correct legal conclusions from multiple options. \textbf{Legal Consultation}~(LCS) presents real legal questions scored by an LLM judge on answer and legal basis quality. From LexEval, we select \textbf{Legal Article Prediction}~(LAP), predicting applicable statutory articles. We exclude purely arithmetic tasks and ensure balanced sampling across temporal contexts.

\paragraph{Legal Article Recitation.}
The seventh task, \textbf{Legal Article Recitation}~(LAR), is our newly constructed task designed to directly probe models' temporal consistency. LAR requires the agent to recite the precise statutory text of a specified article under a given temporal context, evaluated by ROUGE-L~\cite{fei2023lawbenchbenchmarkinglegalknowledge}. We recruit law students from Chinese law schools to annotate provisions that undergo significant textual changes across amendments, covering Criminal Law, Criminal Procedure Law, and Civil Procedure Law across 13 amendment versions from 2000 to 2025. The full annotation guidelines are provided in~\cref{app:lar_annotation}. As shown in~\cref{figs:heatmap}, existing models consistently peak on provisions near their training cutoff (2021--2022) while degrading on both earlier and post-cutoff versions, confirming the temporal bias that motivates our framework.

\subsection{Out-of-Domain Tasks.}
For out-of-domain evaluation, we select from DISC-LawEval~\cite{yue2023disclawllm} a subset of objective multiple-choice questions with clearly identifiable publication dates, spanning six professional examination domains (CPA, PAE, NJE, UNGEE, LBK, PFE). We embed the publication year of each question into its prompt to introduce temporal context, resulting in 768 test instances of which 571 (74\%) carry explicit temporal references.

\section{Methodology}

Our framework addresses the two challenges identified in~\cref{sec:legal_tasks} through a legal toolset for broad and precise retrieval with temporal awareness (\cref{sec:toolset}), and an end-to-end RL training pipeline (\cref{sec:rl_training}). An overview is shown in~\cref{figs:method}.

\subsection{Agentic Legal Toolset}\label{sec:toolset}
We equip the agent with three external tools $\mathcal{T} = \{\texttt{web\_search},\; \texttt{rag\_retrieve},\; \texttt{browse\_webpage}\}$ designed for the dual demands of legal reasoning: broad coverage over diverse legal knowledge and precise, article-level statute retrieval with temporal provenance. Both \texttt{browse\_webpage} and \texttt{rag\_retrieve} employ an auxiliary LLM for content extraction and query analysis, respectively (see~\cref{app:hyperparameters} for model details).

\paragraph{Web Search.}
The web search tool issues queries to an online search API\footnote{\url{https://serper.dev/}} and returns top-$k$ results with title, URL, and snippet. Our tool schema explicitly scopes queries to legal information and provisions not covered by the local RAG corpus, guiding the agent to route statutory queries to RAG while using web search for judicial interpretations, case analyses, and legal commentary.

\paragraph{Browse Webpage.}
The browsing tool fetches target URLs from prior web search results and extracts their content through a legal reading agent with a domain-specific prompt~(\cref{app:tool_prompts}) that enforces verbatim preservation of statutory text, complete extraction of judicial documents, and retention of temporal metadata. A cross-tool context store bridges web search and browsing by maintaining a per-instance log of search results.

\paragraph{Temporal-Enhanced Legal Statute RAG.}
To address the precise statute matching challenge, we construct a temporally-indexed RAG system over a curated corpus of Chinese statutory law. The corpus covers 16 major statutes across all historical amendment versions, with each article chunk annotated with a temporal validity window ($t_{\text{from}}$, $t_{\text{to}}$) indicating the period during which the provision was legally effective. The retrieval pipeline operates in four stages.
First, an LLM-based query analyzer extracts structured information from the legal query (see~\cref{app:tool_prompts} for the full prompt), including temporal references (converted to standardized dates), article/chapter identifiers, and legal keywords. Second, temporal filtering restricts the candidate set to provisions whose validity windows overlap with the extracted temporal context, ensuring that only temporally appropriate statutes are considered. Third, three parallel retrieval channels score the filtered candidates, including keyword exact matching (weight $w_k = 3.0$), dense vector retrieval via FAISS with Chinese legal text embeddings (weight $w_d = 2.0$), and BM25 sparse retrieval (weight $w_s = 1.0$). Finally, Reciprocal Rank Fusion~(RRF) aggregates the three ranked lists into a unified ranking:
\begin{equation}\label{eq:rrf}
\text{score}(d) = \sum_{c \in \{k, d, s\}} \frac{w_c}{K + r_c(d)},
\end{equation}
where $r_c(d)$ is the rank of document $d$ in channel $c$ and $K=60$ is the smoothing constant. The top-$k$ results are returned with the full article text and explicit temporal provenance.
The complementarity of these tools is central to our design: \texttt{rag\_retrieve} ensures precise, article-level statute matching with version-aware retrieval, while \texttt{web\_search} and \texttt{browse\_webpage} provide access to judicial interpretations, case analyses, and legal commentary that lie beyond the statute corpus.

\begin{table*}[t]
\renewcommand{\arraystretch}{1}
\small
\setlength{\tabcolsep}{4pt}
\begin{center}
\begin{tabular}{
    m{2.7cm} 
    m{0.5cm}<{\centering} 
    m{0.2cm}<{\centering} 
    m{0.7cm}<{\centering} 
    |m{0.9cm}<{\centering}|m{0.7cm}<{\centering}
    |m{0.7cm}<{\centering}
    m{0.7cm}<{\centering}m{0.7cm}<{\centering}m{0.7cm}<{\centering}m{0.7cm}<{\centering}
    |m{0.9cm}<{\centering} 
}
\toprule
\multirow{2}{*}{\textbf{Method}} &
\multicolumn{3}{c|}{\textbf{Setting}}&
\textbf{LAR} & \textbf{LCS} & \textbf{KQA} & \textbf{LAP} &  \textbf{CCP} & \textbf{PTP} & \textbf{LCA} & \multirow{2}{*}{\textbf{Avg.}} \\
&
\multirow{1}{*}{\textbf{\scriptsize Think}} &
\multirow{1}{*}{\textbf{\scriptsize Web}} &
\multirow{1}{*}{\textbf{\scriptsize RAG}} &
\multicolumn{1}{c|}{\textbf{\scriptsize Rouge-L}} & \multicolumn{1}{c|}{\textbf{\scriptsize MBE}} & \multicolumn{5}{c|}{\textbf{\scriptsize Accuracy}} \\
\midrule
\multicolumn{12}{l}{\textbf{\textit{$\bullet$ Legal LLMs}}} \\
DISC-LawLLM       & \color{red!70!black}\ding{55} & \color{red!70!black}\ding{55} & \color{red!70!black}\ding{55} & 37.41 & 7.81 & 14.84 & 46.88 & 54.69 & 7.81 &  27.34 &28.11\\
DISC-LawLLM       & \color{green!70!black}\ding{52} & \color{red!70!black}\ding{55} & \color{red!70!black}\ding{55} &35.37 & 8.59 & 25.00 & 39.84 &  53.12 &0.00  & 19.53 & 25.92 \\
LegalDelta        & \color{red!70!black}\ding{55} & \color{red!70!black}\ding{55} & \color{red!70!black}\ding{55} &52.30  & 24.61 & 47.66 & 81.25 & 60.94 & 20.31 &  53.91 & 48.71 \\
LegalDelta        & \color{green!70!black}\ding{52} & \color{red!70!black}\ding{55} & \color{red!70!black}\ding{55} &52.31  & 26.17 & 45.31 & 81.25 & 60.16 & 17.97 & 54.69 & 48.27  \\
\midrule
\multicolumn{12}{l}{$\spadesuit$ \textbf{\textit{Vanilla LLMs}}} \\
GPT-4o-mini        & \color{red!70!black}\ding{55} & \color{red!70!black}\ding{55} & \color{red!70!black}\ding{55} & 17.15  &  7.42  &  29.69 & 13.28 & 19.53 & 10.94 & 16.41 & 16.35 \\
\rowcolor{pink!30}
Qwen-2.5-7b        & \color{red!70!black}\ding{55} & \color{red!70!black}\ding{55} & \color{red!70!black}\ding{55} & 36.39  &  14.84  &  35.94 & 68.75 & 40.62 & 9.38 & 31.25  &  33.88\\
Qwen-3-8b        & \color{red!70!black}\ding{55} & \color{red!70!black}\ding{55} & \color{red!70!black}\ding{55} & 33.53  &  28.52  &  45.31 & 78.12 & 37.50 & 17.19 & 46.88  &  41.01\\
Qwen-3-14b        & \color{red!70!black}\ding{55} & \color{red!70!black}\ding{55} & \color{red!70!black}\ding{55}  & 38.74  &  33.59  &  52.34 & 82.81 & 46.88 & 8.59 & 50.78  &  44.82 \\
Qwen-3-30b-A3b        & \color{red!70!black}\ding{55} & \color{red!70!black}\ding{55} & \color{red!70!black}\ding{55} & 51.27  &  41.41  &  55.47 & 85.16 & 42.19 & 10.16 & 48.44  &  47.73\\
\midrule
\multicolumn{12}{l}{$\clubsuit$ \textbf{\textit{LRMs}}} \\
o4-mini & \color{green!70!black}\ding{52} & \color{red!70!black}\ding{55} & \color{red!70!black}\ding{55}  & 19.92  &  18.75  &  37.50 & 57.03 & 34.38 & 10.16 & 33.59  &  30.19 \\
QwQ-32b & \color{green!70!black}\ding{52} & \color{red!70!black}\ding{55} & \color{red!70!black}\ding{55}  & 39.64  &  38.67  &  55.47 & 53.91 & 37.50 & 10.16 & 53.91  &  41.32 \\
Qwen-3-30b-A3b & \color{green!70!black}\ding{52} & \color{red!70!black}\ding{55} & \color{red!70!black}\ding{55} & 43.14  &  37.11  &  50.78 & 82.81 & 46.09 & 12.50 & 53.91  &  46.62 \\
DeepSeek-V3.2 & \color{green!70!black}\ding{52} & \color{red!70!black}\ding{55} & \color{red!70!black}\ding{55} & 63.30  &  42.97  &  62.50 & 78.12 & 43.75 & 13.28 & 50.00  &  50.56 \\
\midrule
\multicolumn{12}{l}{$\diamond$ \textbf{\textit{Deep Research}}} \\
Search-r1-base      & \color{green!70!black}\ding{52} & \color{red!70!black}\ding{55} & \color{green!70!black}\ding{52} & 21.38 & 16.02 & 43.75 &  64.06& 22.66 & 3.91 & 36.72 &29.78 \\
Search-r1-instruct & \color{green!70!black}\ding{52} & \color{red!70!black}\ding{55} & \color{green!70!black}\ding{52} & 35.31 & 19.53 & 32.81  & 65.62  & 34.38 & 14.84 & 39.06 & 34.51 \\
DeepResearcher    & \color{green!70!black}\ding{52} & \color{green!70!black}\ding{52} & \color{red!70!black}\ding{55} & 40.51 & 18.75 & 39.06 & 62.50 & 38.28 & 13.28 & 36.72 & 35.59 \\
DeepPlanner    & \color{green!70!black}\ding{52} & \color{green!70!black}\ding{52} & \color{red!70!black}\ding{55} & 54.02  &  20.31  &  39.84 & 75.00 & 36.72 & 13.28 & 35.16  &  39.19  \\
\midrule
\rowcolor{cyan!15}
\textbf{LegalSearch-R1}    & \color{green!70!black}\ding{52} & \color{green!70!black}\ding{52} & \color{green!70!black}\ding{52} & 96.73  &  35.16  &  50.00 & 87.50 & 62.50 & 13.28 & 46.09  &  55.90\\
\bottomrule
\end{tabular}
\end{center}
\vspace{-0.1in}
\caption{In-domain evaluation results across seven legal tasks. \colorbox{pink!30}{Pink} highlights the base model of \papertitle.}
\vspace{-0.15in}
\label{tabs:evaluation_overall_performance}
\end{table*}

\subsection{Agentic RL Training}\label{sec:rl_training}

We train the agent end-to-end via GRPO~\cite{shao2024deepseekmathpushinglimitsmathematical} with entropy-based advantage shaping~\cite{fan2025deepplannerscalingplanningcapability} on temporally-indexed data spanning 13 amendment versions.

\paragraph{GRPO with Entropy-Based Advantage Shaping.}
We use token-level GRPO~\cite{yu2025dapoopensourcellmreinforcement} to preserve individual token contributions in long legal search trajectories. For each query $(q \mid t_q)$, the agent generates $G$ rollouts. The objective is
\begin{equation}\label{eq:grpo}
\small
\mathcal{J}(\theta) \!=\! \sum_{i,t}
    \big[ \min \big( r_{i,t}\, A_{i,t},\; \text{clip}(r_{i,t}, 1{\pm}\epsilon)\, A_{i,t} \big)
    \!-\! \beta\, D_{\text{KL}} \big],
\end{equation}
normalized by total agent-generated tokens $\sum_{i=1}^G |y_i|$ (excluding tool responses), where $r_{i,t} = {\pi_{\theta}(a_{i,t} \mid s, a_{i,<t})}/{\pi_{\text{old}}(a_{i,t} \mid s, a_{i,<t})}$. The group-relative advantage $A_{i} = ({R_i - \text{mean}(\mathbf{R})})/{\text{std}(\mathbf{R})}$ is computed over rollout rewards within each group.
In legal search, the agent must decide during planning whether to incorporate temporal constraints into its search queries, a high-entropy decision that determines whether the correct statute version is retrieved. Following~\citet{fan2025deepplannerscalingplanningcapability}, we add a gradient-detached shaping term to amplify advantage signals for these planning tokens, accelerating the emergence of temporal query formulation while preventing entropy collapse:
\begin{equation}\label{eq:shaping}
\begin{aligned}
\small
\psi(\mathcal{H}_{i,t}) &= \min\left(
    {\alpha} \cdot {\mathcal{H}_{i,t}^{\text{detach}}}, \;
    \frac{|A_{i,t}|}{{\kappa}}
\right),\\
A^{\text{EAS}}_{i,t} &= A_{i,t} + \psi(\mathcal{H}_{i,t}),
\end{aligned}
\end{equation}
where $\mathcal{H}_{i,t}$ is the per-token entropy (detached from the computation graph), $\alpha$ is the shaping coefficient, and $\kappa > 1$ prevents flipping negative advantages to positive.

\paragraph{Reward Function.}
The reward $R$ balances structural compliance with task-specific correctness across heterogeneous legal answer forms:
\begin{equation}\label{eq:reward}
\small
R = \begin{cases}
0, & \text{if } \textsc{Format}(\tau) = \texttt{false},\\
\textsc{Score}(a, a^*, \text{task}), & \text{otherwise},
\end{cases}
\end{equation}
where $\textsc{Format}(\tau)$ verifies balanced XML tags and a plan in the first round. The task-specific scoring is
\begin{equation}\label{eq:score}
\small
\textsc{Score} = \begin{cases}
\mathds{1}[\text{ROUGE-L}(a, a^*) \geq 0.95], & \text{LAR},\\
\mathds{1}[\text{Match}_{\text{unord}}(a, a^*)], & \text{CCP},\\
\mathds{1}[\text{LLM-Judge}(a, a^*)], & \text{LCS},\\
\mathds{1}[a = a^*], & \text{otherwise},
\end{cases}
\end{equation}
where $a$ is the predicted answer and $a^*$ the ground truth. LAR uses character-level ROUGE-L with a strict threshold, CCP employs unordered matching with suffix normalization, and LCS uses an LLM judge evaluating both answer correctness and legal basis quality. All scores are binary.

\section{Experiments}\label{sec:experiments}

\subsection{Baselines}
We compare \papertitle against four categories of baselines. (1)~\textbf{Legal LLMs} include DISC-LawLLM~\cite{yue2023disclawllm} and LegalDelta~\cite{dai2025legaldelta}, both evaluated with and without chain-of-thought reasoning. (2)~\textbf{Vanilla LLMs} include GPT-4o-mini, our backbone Qwen-2.5-7B~\cite{qwen2025qwen25technicalreport}, and the Qwen-3 series (8B, 14B, 30B-A3B)~\cite{qwen2025qwen3technicalreport}. (3)~\textbf{Large Reasoning Models~(LRMs)} include o4-mini, QwQ-32B, Qwen-3-30B-A3B with thinking, and DeepSeek-V3.2~\cite{deepseekai2025deepseekv3technicalreport}. (4)~\textbf{Deep Research} agents include Search-R1~\cite{jin2025searchr1trainingllmsreason} (base and instruct), DeepResearcher~\cite{zheng2025deepresearcherscalingdeepresearch}, and DeepPlanner~\cite{fan2025deepplannerscalingplanningcapability}. All deep research baselines share the same backbone (Qwen-2.5-7B-Instruct) and are trained on our legal data for fair comparison. Detailed settings for LegalSearch-R1 are provided in~\cref{app:hyperparameters}.

\subsection{Datasets and Metrics}
We evaluate on the benchmark described in~\cref{sec:legal_tasks}. The training set contains 3{,}584 samples (512 per task across all 7 in-domain tasks), the in-domain test set contains 896 samples (128 per task), and the out-of-domain test set contains 768 samples (128 per task across 6 professional examination tasks).
For evaluation, LAR is scored by ROUGE-L (character-level tokenization), and LCS is scored by an LLM judge that evaluates answer correctness and legal basis quality. KQA, LAP, CCP, PTP, and LCA are evaluated by accuracy. All OOD tasks use accuracy. Note that training rewards differ from evaluation metrics for two tasks. During training, LAR uses a binary ROUGE-L threshold ($\geq 0.95$) to provide a clear learning signal, while at evaluation we report the continuous ROUGE-L score. Similarly, LCS training rewards require both answer and legal basis to be judged correct for a binary reward of 1, while at evaluation we report the mean of the two judge scores.

\begin{figure}[t!]
    \centering
    \includegraphics[width=0.47\textwidth]{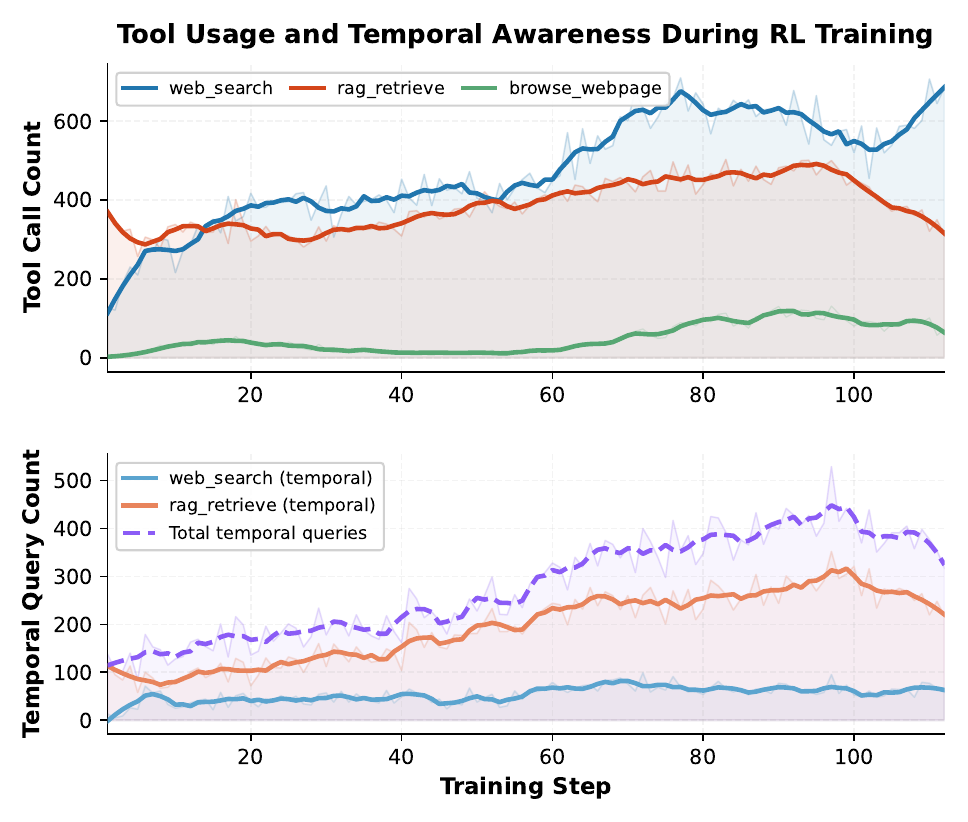}
    \caption{Tool usage and temporal awareness during RL training. Top: tool call counts. Bottom: queries with explicit temporal constraints.}
    \vspace{-0.15in}
    \label{figs:training_tool_usage}
\end{figure}

\subsection{Main Results}
\paragraph{In-Domain Performance.}
As shown in \cref{tabs:evaluation_overall_performance}, LegalSearch-R1 achieves an average score of 55.90 across 7 in-domain tasks, outperforming the strongest deep research baseline DeepPlanner (39.19) by 42.7\% and the strongest LRM DeepSeek-V3.2 (50.56) by 10.6\%. Among legal-specialized LLMs, LegalDelta achieves 48.71 without thinking and 48.27 with thinking, while DISC-LawLLM scores below 30 in both settings. Vanilla LLMs show a clear scaling trend from Qwen-2.5-7B (33.88) to Qwen-3-30B-A3B (47.73), yet none matches our 7B agent.
The most striking result appears on the LAR task, where LegalSearch-R1 achieves 96.73 ROUGE-L, surpassing DeepSeek-V3.2 (63.30) by over 33 points and all deep research agents by more than 42 points. This dramatic gap directly validates the effectiveness of our local statute RAG for precise article-level retrieval, as all baselines relying on parametric knowledge or web search alone fail to recite statutory text with comparable fidelity.
Excluding LAR, our agent still outperforms the best deep research baseline DeepPlanner by 33.7\% and remains competitive with DeepSeek-V3.2 despite using significantly fewer parameters.

\paragraph{Out-of-Domain Performance.}
On 6 out-of-domain professional examination tasks (\cref{app:ood_results}), LegalSearch-R1 achieves 63.67, outperforming all deep research baselines (best 53.91 from DeepPlanner) and all legal LLMs (best 49.35 from LegalDelta). Our 7B agent is competitive with Qwen-3-30B-A3B (63.80) and DeepSeek-V3.2 (66.41), models with 4$\times$ to 30$\times$ more parameters. This robust generalization demonstrates that the temporally-indexed training and dual-tool architecture transfer effectively to unseen legal domains and examination formats.

\subsection{Temporal Consistency Analysis}

We evaluate whether RL training on temporally-indexed data teaches the agent to proactively incorporate temporal constraints into its search behavior through two complementary analyses.
As shown in~\cref{figs:temporal_query}, we count search queries that include explicit temporal references across all deep research frameworks. Baseline agents produce between 97 and 227 temporally-aware queries, while LegalSearch-R1 generates 502 in-domain and 537 out-of-domain temporal queries, a 121\% to 454\% increase. This gap emerges entirely from RL training, as all agents share the same backbone and system prompt.
\cref{figs:training_tool_usage} traces tool usage and temporal query counts across 112 training steps. The bottom panel reveals a clear upward trend in temporal queries, rising from approximately 100 to over 250, with web search temporal queries showing the most pronounced growth. This confirms that the agent progressively learns to embed temporal constraints through outcome-based RL, without explicit supervision on query formulation.
Combined with the heatmap analysis in~\cref{figs:heatmap}, which shows that baselines suffer from temporal bias anchored to their training cutoff, these results demonstrate that LegalSearch-R1 achieves 57.7\% to 80.3\% improvement on temporal consistency by learning to act on the temporal context of each query.

\subsection{Effect of Temporal-Enhanced RAG}

\begin{figure}[t!]
    \centering
    \includegraphics[width=0.5\textwidth]{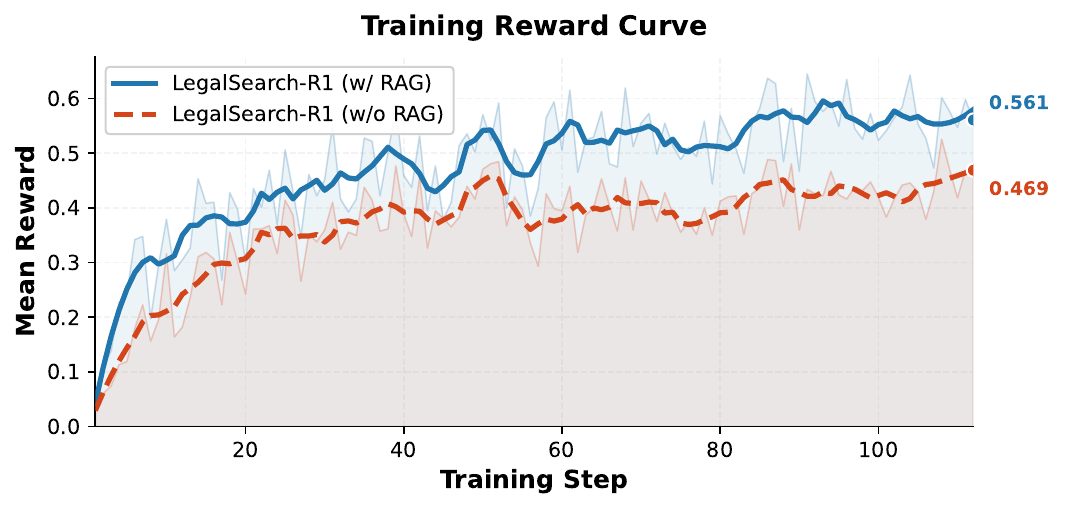}
    \caption{Training reward curves with and without the local temporal-enhanced RAG module.}
    \vspace{-0.1in}
    \label{figs:training_reward_curve}
\end{figure}

To isolate the contribution of the local statute RAG module, we train a variant of LegalSearch-R1 that uses only \texttt{web\_search} and \texttt{browse\_webpage} without \texttt{rag\_retrieve}, keeping all other settings identical. As shown in~\cref{figs:training_reward_curve}, the variant with RAG achieves consistently higher training rewards than the variant without, reaching 0.561 versus 0.469 at convergence.
The impact is most pronounced on the LAR task, where our full model achieves 96.73 ROUGE-L while all baselines without structured statute retrieval remain below 63.30. This confirms that web search alone cannot provide the precise, article-level statutory text that legal reasoning demands. The local RAG corpus, with its temporally-indexed provisions and hybrid retrieval pipeline, fills a critical gap that no amount of web search can bridge, particularly for tasks requiring verbatim statute recitation.

\subsection{Case Study}

A detailed case in~\cref{app:case_study} illustrates the hybrid search architecture in practice. In an inheritance dispute (2001--2004), the agent first uses \texttt{web\_search} to discover that the priority of notarized wills changed with the 2021 Civil Code, then calls \texttt{rag\_retrieve} to retrieve the precise statutory text. Crucially, the temporally-enhanced RAG filters retrieved statutes by validity period: given the case timeframe, it returns the pre-amendment Inheritance Law rather than the post-2021 Civil Code provisions that would otherwise dominate retrieval results. The agent thus correctly applies the old Inheritance Law (Art.~20), under which audio-recorded wills cannot override notarized wills, arriving at the correct answer. Without temporal filtering, baselines uniformly apply the current Civil Code and reach the opposite conclusion.

\section{Conclusion}
We identify temporal inconsistency and inadequate precise statute matching as two critical yet overlooked challenges in legal agentic search. To address them, we propose \papertitle, an end-to-end RL framework pairing temporally-indexed training with a mixed architecture of local statute RAG and online web search, enhanced by entropy-based advantage shaping to accelerate temporal query planning. Experimental results demonstrate that our agent significantly outperforms state-of-the-art baselines, with strong temporal consistency and robust out-of-domain generalization.

\newpage

\section*{Limitations}
Our research is conducted exclusively within the Chinese legal system, utilizing Chinese-language statutes and legal tasks. This may limit generalizability to other legal systems, particularly common law jurisdictions where legal reasoning relies more on judicial precedents than codified statutes, and multilingual performance remains unassessed.
Regarding baseline selection, we exclude LRAS~\cite{zhou2026lras} because its model weights are not publicly released. We also omit several open-source legal LLMs such as Fuzi-Mingcha~\cite{wu2023fuzimingcha} and Hanfei~\cite{he2023hanfei} because these models consistently fail to follow our evaluation protocol, producing outputs that cannot be parsed into the required \texttt{<answer>}$\cdots$\texttt{</answer>} format.
Furthermore, our RAG corpus indexes only statutory provisions and does not incorporate judicial precedents or case law. This reflects the civil law tradition of the Chinese legal system, where codified statutes serve as the primary source of law. Incorporating case law would introduce challenges in structuring the heterogeneous content of judicial opinions for precise retrieval, and we leave this extension to future work.

\section*{Ethics Statement}
All benchmark tasks in this work are derived from publicly available datasets, including LawBench~\cite{fei2023lawbenchbenchmarkinglegalknowledge}, LexEval~\cite{li2024lexeval}, and DISC-LawEval~\cite{yue2023disclawllm}, which are released under open-source licenses for academic research. The Legal Article Recitation~(LAR) task is annotated by law students from Chinese law schools, and all annotators are compensated at a rate of 15 USD per hour, which exceeds the prevailing local minimum wage. The annotated data contains only statutory provisions from publicly promulgated laws and does not involve any personally identifiable information. We acknowledge that legal AI systems carry inherent risks of misapplication in high-stakes settings, and our framework and benchmark are released solely for academic research purposes. Our experiments employ search tools\footnote{\url{https://serper.dev/}} and LLM APIs\footnote{\url{https://openrouter.ai/}} under their respective licenses, and we adhered to the terms of service for all APIs used.
\bibliography{custom}

@misc{guha2023legalbenchcollaborativelybuiltbenchmark,
      title={LegalBench: A Collaboratively Built Benchmark for Measuring Legal Reasoning in Large Language Models}, 
      author={Neel Guha and Julian Nyarko and Daniel E. Ho and Christopher Ré and Adam Chilton and Aditya Narayana and Alex Chohlas-Wood and Austin Peters and Brandon Waldon and Daniel N. Rockmore and Diego Zambrano and Dmitry Talisman and Enam Hoque and Faiz Surani and Frank Fagan and Galit Sarfaty and Gregory M. Dickinson and Haggai Porat and Jason Hegland and Jessica Wu and Joe Nudell and Joel Niklaus and John Nay and Jonathan H. Choi and Kevin Tobia and Margaret Hagan and Megan Ma and Michael Livermore and Nikon Rasumov-Rahe and Nils Holzenberger and Noam Kolt and Peter Henderson and Sean Rehaag and Sharad Goel and Shang Gao and Spencer Williams and Sunny Gandhi and Tom Zur and Varun Iyer and Zehua Li},
      year={2023},
      eprint={2308.11462},
      archivePrefix={arXiv},
      primaryClass={cs.CL},
      url={https://arxiv.org/abs/2308.11462}, 
}

@misc{fei2023lawbenchbenchmarkinglegalknowledge,
      title={LawBench: Benchmarking Legal Knowledge of Large Language Models}, 
      author={Zhiwei Fei and Xiaoyu Shen and Dawei Zhu and Fengzhe Zhou and Zhuo Han and Songyang Zhang and Kai Chen and Zongwen Shen and Jidong Ge},
      year={2023},
      eprint={2309.16289},
      archivePrefix={arXiv},
      primaryClass={cs.CL},
      url={https://arxiv.org/abs/2309.16289}, 
}

@misc{openai2024gpt4technicalreport,
      title={GPT-4 Technical Report}, 
      author={Achiam, Josh and Adler, Steven and Agarwal, Sandhini and Ahmad, Lama and Akkaya, Ilge and Aleman, Florencia Leoni and Almeida, Diogo and Altenschmidt, Janko and Altman, Sam and Anadkat, Shyamal and others},
      year={2024},
      eprint={2303.08774},
      archivePrefix={arXiv},
      primaryClass={cs.CL},
      url={https://arxiv.org/abs/2303.08774}, 
}

@misc{qwen2025qwen25technicalreport,
      title={Qwen2.5 Technical Report}, 
      author={Qwen and : and An Yang and Baosong Yang and Beichen Zhang and Binyuan Hui and Bo Zheng and Bowen Yu and Chengyuan Li and Dayiheng Liu and Fei Huang and Haoran Wei and Huan Lin and Jian Yang and Jianhong Tu and Jianwei Zhang and Jianxin Yang and Jiaxi Yang and Jingren Zhou and Junyang Lin and Kai Dang and Keming Lu and Keqin Bao and Kexin Yang and Le Yu and Mei Li and Mingfeng Xue and Pei Zhang and Qin Zhu and Rui Men and Runji Lin and Tianhao Li and Tianyi Tang and Tingyu Xia and Xingzhang Ren and Xuancheng Ren and Yang Fan and Yang Su and Yichang Zhang and Yu Wan and Yuqiong Liu and Zeyu Cui and Zhenru Zhang and Zihan Qiu},
      year={2025},
      eprint={2412.15115},
      archivePrefix={arXiv},
      primaryClass={cs.CL},
      url={https://arxiv.org/abs/2412.15115}, 
}

@misc{qwen2025qwen3technicalreport,
      title={Qwen3 Technical Report}, 
      author={An Yang and Anfeng Li and Baosong Yang and Beichen Zhang and Binyuan Hui and Bo Zheng and Bowen Yu and Chang Gao and Chengen Huang and Chenxu Lv and Chujie Zheng and Dayiheng Liu and Fan Zhou and Fei Huang and Feng Hu and Hao Ge and Haoran Wei and Huan Lin and Jialong Tang and Jian Yang and Jianhong Tu and Jianwei Zhang and Jianxin Yang and Jiaxi Yang and Jing Zhou and Jingren Zhou and Junyang Lin and Kai Dang and Keqin Bao and Kexin Yang and Le Yu and Lianghao Deng and Mei Li and Mingfeng Xue and Mingze Li and Pei Zhang and Peng Wang and Qin Zhu and Rui Men and Ruize Gao and Shixuan Liu and Shuang Luo and Tianhao Li and Tianyi Tang and Wenbiao Yin and Xingzhang Ren and Xinyu Wang and Xinyu Zhang and Xuancheng Ren and Yang Fan and Yang Su and Yichang Zhang and Yinger Zhang and Yu Wan and Yuqiong Liu and Zekun Wang and Zeyu Cui and Zhenru Zhang and Zhipeng Zhou and Zihan Qiu},
      year={2025},
      eprint={2505.09388},
      archivePrefix={arXiv},
      primaryClass={cs.CL},
      url={https://arxiv.org/abs/2505.09388}, 
}

@software{wu2023fuzimingcha,
  title={Fuzi.Mingcha},
  author={Wu, Shiguang and Liu, Zhongkun and Zhang, Zhen and Chen, Zheng and Deng, Wentao and Zhang, Wenhao and Yang, Jiyuan and Yao, Zhitao and Lyu, Yougang and Xin, Xin and Gao, Shen and Ren, Pengjie and Ren, Zhaochun and Chen, Zhumin},
  year={2023},
  publisher={GitHub},
  howpublished={\url{https://github.com/irlab-sdu/fuzi.mingcha}},
}

@misc{deepseekai2025deepseekv3technicalreport,
      title={DeepSeek-V3 Technical Report}, 
      author={DeepSeek-AI and Aixin Liu and Bei Feng and Bing Xue and Bingxuan Wang and others},
      year={2025},
      eprint={2412.19437},
      archivePrefix={arXiv},
      primaryClass={cs.CL},
      url={https://arxiv.org/abs/2412.19437}, 
}

@misc{cui2023chatlaw,
      title={Chatlaw: A Multi-Agent Collaborative Legal Assistant with Knowledge Graph Enhanced Mixture-of-Experts Large Language Model}, 
      author={Jiaxi Cui and Munan Ning and Zongjian Li and Bohua Chen and Yang Yan and Hao Li and Bin Ling and Yonghong Tian and Li Yuan},
      year={2024},
      eprint={2306.16092},
      archivePrefix={arXiv},
      primaryClass={cs.CL},
      url={https://arxiv.org/abs/2306.16092}, 
}

@article{yue2023disclawllm,
    author = {Shengbin Yue and Wei Chen and Siyuan Wang and Bingxuan Li and Chenchen Shen and Shujun Liu and Yuxuan Zhou and Yao Xiao and Song Yun and Xuanjing Huang and Zhongyu Wei},
    title = {{DISC-LawLLM:} Fine-tuning Large Language Models for Intelligent Legal Services},
    journal = {CoRR},
    volume = {abs/2309.11325},
    year = {2023},
    url = {https://doi.org/10.48550/arXiv.2309.11325},
    doi = {10.48550/ARXIV.2309.11325},
    eprinttype = {arXiv},
    eprint = {2309.11325},
}

@misc{dai2025legaldelta,
      title={Legal$\Delta$: Enhancing Legal Reasoning in LLMs via Reinforcement Learning with Chain-of-Thought Guided Information Gain}, 
      author={Xin Dai and Buqiang Xu and Zhenghao Liu and Yukun Yan and Huiyuan Xie and Xiaoyuan Yi and Shuo Wang and Ge Yu},
      year={2026},
      eprint={2508.12281},
      archivePrefix={arXiv},
      primaryClass={cs.CL},
      url={https://arxiv.org/abs/2508.12281}, 
}

@inproceedings{zhong2020iteratively,
  title={Iteratively Questioning and Answering for Interpretable Legal Judgment Prediction},
  author={Haoxiang Zhong and Yuzhong Wang and Cunchao Tu and T. Zhang and Zhiyuan Liu and Maosong Sun},
  booktitle={AAAI Conference on Artificial Intelligence},
  year={2020},
  url={https://api.semanticscholar.org/CorpusID:213738448}
}

@inproceedings{louis2024interpretable,
      title={Interpretable Long-Form Legal Question Answering with Retrieval-Augmented Large Language Models},
      author={Antoine Louis and Gijs van Dijck and Gerasimos Spanakis},
      year={2024},
      booktitle={Proceedings of the 38th AAAI Conference on Artificial Intelligence},
      pages={18642--18650},
      publisher={AAAI Press},
      doi={10.1609/aaai.v38i20.30232},
      eprint={2309.17050},
      archivePrefix={arXiv},
      primaryClass={cs.CL},
      url={https://ojs.aaai.org/index.php/AAAI/article/view/30232},
}

@inproceedings{fan-etal-2024-goldcoin,
    title = "{G}old{C}oin: Grounding Large Language Models in Privacy Laws via Contextual Integrity Theory",
    author = "Fan, Wei  and
      Li, Haoran  and
      Deng, Zheye  and
      Wang, Weiqi  and
      Song, Yangqiu",
    editor = "Al-Onaizan, Yaser  and
      Bansal, Mohit  and
      Chen, Yun-Nung",
    booktitle = "Proceedings of the 2024 Conference on Empirical Methods in Natural Language Processing",
    month = nov,
    year = "2024",
    address = "Miami, Florida, USA",
    publisher = "Association for Computational Linguistics",
    url = "https://aclanthology.org/2024.emnlp-main.195/",
    doi = "10.18653/v1/2024.emnlp-main.195",
    pages = "3321--3343"
}

@misc{lewis2021retrievalaugmentedgenerationknowledgeintensivenlp,
      title={Retrieval-Augmented Generation for Knowledge-Intensive NLP Tasks}, 
      author={Patrick Lewis and Ethan Perez and Aleksandra Piktus and Fabio Petroni and Vladimir Karpukhin and Naman Goyal and Heinrich Küttler and Mike Lewis and Wen-tau Yih and Tim Rocktäschel and Sebastian Riedel and Douwe Kiela},
      year={2021},
      eprint={2005.11401},
      archivePrefix={arXiv},
      primaryClass={cs.CL},
      url={https://arxiv.org/abs/2005.11401}, 
}

@misc{gao2024retrievalaugmentedgenerationlargelanguage,
      title={Retrieval-Augmented Generation for Large Language Models: A Survey}, 
      author={Yunfan Gao and Yun Xiong and Xinyu Gao and Kangxiang Jia and Jinliu Pan and Yuxi Bi and Yi Dai and Jiawei Sun and Meng Wang and Haofen Wang},
      year={2024},
      eprint={2312.10997},
      archivePrefix={arXiv},
      primaryClass={cs.CL},
      url={https://arxiv.org/abs/2312.10997}, 
}

@misc{openaideepresearch,
  title={Deep research System Card},
  author = {OpenAI},
  year = {2025},
  url={https://openai.com/index/deep-research-system-card/}
}

@misc{geminideepresearch,
  title={Gemini Deep Research},
  author = {Gemini},
  year = {2025},
  url={https://gemini.google/overview/deep-research/}
}

@misc{zheng2025deepresearcherscalingdeepresearch,
      title={DeepResearcher: Scaling Deep Research via Reinforcement Learning in Real-world Environments}, 
      author={Yuxiang Zheng and Dayuan Fu and Xiangkun Hu and Xiaojie Cai and Lyumanshan Ye and Pengrui Lu and Pengfei Liu},
      year={2025},
      eprint={2504.03160},
      archivePrefix={arXiv},
      primaryClass={cs.AI},
      url={https://arxiv.org/abs/2504.03160}, 
}

@misc{deepseekai2025deepseekr1incentivizingreasoningcapability,
      title={DeepSeek-R1: Incentivizing Reasoning Capability in LLMs via Reinforcement Learning}, 
      author={DeepSeek-AI and Daya Guo and Dejian Yang and Haowei Zhang and Junxiao Song and others},
      year={2025},
      eprint={2501.12948},
      archivePrefix={arXiv},
      primaryClass={cs.CL},
      url={https://arxiv.org/abs/2501.12948}, 
}

@misc{openai2026openaio1card,
      title={OpenAI o1 System Card}, 
      author={OpenAI and : and Aaron Jaech and Adam Kalai and Adam Lerer and Adam Richardson and Ahmed El-Kishky and Aiden Low and Alec Helyar and Aleksander Madry and Alex Beutel and Alex Carney and Alex Iftimie and Alex Karpenko and Alex Tachard Passos and Alexander Neitz and Alexander Prokofiev and Alexander Wei and Allison Tam and Ally Bennett and Ananya Kumar and Andre Saraiva and Andrea Vallone and Andrew Duberstein and Andrew Kondrich and Andrey Mishchenko and Andy Applebaum and Angela Jiang and Ashvin Nair and Barret Zoph and Behrooz Ghorbani and Bohan Zhang and Ben Rossen and Benjamin Sokolowsky and Boaz Barak and Bob McGrew and Borys Minaiev and Botao Hao and Bowen Baker and Brandon Houghton and Brandon McKinzie and Brydon Eastman and Camillo Lugaresi and Cary Bassin and Cary Hudson and Chak Ming Li and Charles de Bourcy and Chelsea Voss and Chen Shen and Chong Zhang and Chris Koch and Chris Orsinger and Christopher Hesse and Claudia Fischer and Clive Chan and Dan Roberts and Daniel Kappler and Daniel Levy and Daniel Selsam and David Dohan and David Farhi and David Mely and David Robinson and Dimitris Tsipras and Doug Li and Dragos Oprica and Eben Freeman and Eddie Zhang and Edmund Wong and Elizabeth Proehl and Enoch Cheung and Eric Mitchell and Eric Wallace and Erik Ritter and Evan Mays and Fan Wang and Felipe Petroski Such and Filippo Raso and Florencia Leoni and Foivos Tsimpourlas and Francis Song and Fred von Lohmann and Freddie Sulit and Geoff Salmon and Giambattista Parascandolo and Gildas Chabot and Grace Zhao and Greg Brockman and Guillaume Leclerc and Hadi Salman and Haiming Bao and Hao Sheng and Hart Andrin and Hessam Bagherinezhad and Hongyu Ren and Hunter Lightman and Hyung Won Chung and Ian Kivlichan and Ian O'Connell and Ian Osband and Ignasi Clavera Gilaberte and Ilge Akkaya and Ilya Kostrikov and Ilya Sutskever and Irina Kofman and Jakub Pachocki and James Lennon and Jason Wei and Jean Harb and Jerry Twore and Jiacheng Feng and Jiahui Yu and Jiayi Weng and Jie Tang and Jieqi Yu and Joaquin Quiñonero Candela and Joe Palermo and Joel Parish and Johannes Heidecke and John Hallman and John Rizzo and Jonathan Gordon and Jonathan Uesato and Jonathan Ward and Joost Huizinga and Julie Wang and Kai Chen and Kai Xiao and Karan Singhal and Karina Nguyen and Karl Cobbe and Katy Shi and Kayla Wood and Kendra Rimbach and Keren Gu-Lemberg and Kevin Liu and Kevin Lu and Kevin Stone and Kevin Yu and Lama Ahmad and Lauren Yang and Leo Liu and Leon Maksin and Leyton Ho and Liam Fedus and Lilian Weng and Linden Li and Lindsay McCallum and Lindsey Held and Lorenz Kuhn and Lukas Kondraciuk and Lukasz Kaiser and Luke Metz and Madelaine Boyd and Maja Trebacz and Manas Joglekar and Mark Chen and Marko Tintor and Mason Meyer and Matt Jones and Matt Kaufer and Max Schwarzer and Meghan Shah and Mehmet Yatbaz and Melody Y. Guan and Mengyuan Xu and Mengyuan Yan and Mia Glaese and Mianna Chen and Michael Lampe and Michael Malek and Michele Wang and Michelle Fradin and Mike McClay and Mikhail Pavlov and Miles Wang and Mingxuan Wang and Mira Murati and Mo Bavarian and Mostafa Rohaninejad and Nat McAleese and Neil Chowdhury and Neil Chowdhury and Nick Ryder and Nikolas Tezak and Noam Brown and Ofir Nachum and Oleg Boiko and Oleg Murk and Olivia Watkins and Patrick Chao and Paul Ashbourne and Pavel Izmailov and Peter Zhokhov and Rachel Dias and Rahul Arora and Randall Lin and Rapha Gontijo Lopes and Raz Gaon and Reah Miyara and Reimar Leike and Renny Hwang and Rhythm Garg and Robin Brown and Roshan James and Rui Shu and Ryan Cheu and Ryan Greene and Saachi Jain and Sam Altman and Sam Toizer and Sam Toyer and Samuel Miserendino and Sandhini Agarwal and Santiago Hernandez and Sasha Baker and Scott McKinney and Scottie Yan and Shengjia Zhao and Shengli Hu and Shibani Santurkar and Shraman Ray Chaudhuri and Shuyuan Zhang and Siyuan Fu and Spencer Papay and Steph Lin and Suchir Balaji and Suvansh Sanjeev and Szymon Sidor and Tal Broda and Aidan Clark and Tao Wang and Taylor Gordon and Ted Sanders and Tejal Patwardhan and Thibault Sottiaux and Thomas Degry and Thomas Dimson and Tianhao Zheng and Timur Garipov and Tom Stasi and Trapit Bansal and Trevor Creech and Troy Peterson and Tyna Eloundou and Valerie Qi and Vineet Kosaraju and Vinnie Monaco and Vitchyr Pong and Vlad Fomenko and Weiyi Zheng and Wenda Zhou and Wenting Zhan and Wes McCabe and Wojciech Zaremba and Yann Dubois and Yinghai Lu and Yining Chen and Young Cha and Yu Bai and Yuchen He and Yuchen Zhang and Yunyun Wang and Zheng Shao and Zhuohan Li},
      year={2026},
      eprint={2412.16720},
      archivePrefix={arXiv},
      primaryClass={cs.AI},
      url={https://arxiv.org/abs/2412.16720}, 
}

@misc{yao2023reactsynergizingreasoningacting,
      title={ReAct: Synergizing Reasoning and Acting in Language Models}, 
      author={Shunyu Yao and Jeffrey Zhao and Dian Yu and Nan Du and Izhak Shafran and Karthik Narasimhan and Yuan Cao},
      year={2023},
      eprint={2210.03629},
      archivePrefix={arXiv},
      primaryClass={cs.CL},
      url={https://arxiv.org/abs/2210.03629}, 
}

@misc{jin2025searchr1trainingllmsreason,
      title={Search-R1: Training LLMs to Reason and Leverage Search Engines with Reinforcement Learning}, 
      author={Bowen Jin and Hansi Zeng and Zhenrui Yue and Jinsung Yoon and Sercan Arik and Dong Wang and Hamed Zamani and Jiawei Han},
      year={2025},
      eprint={2503.09516},
      archivePrefix={arXiv},
      primaryClass={cs.CL},
      url={https://arxiv.org/abs/2503.09516}, 
}

@misc{song2025r1searcherincentivizingsearchcapability,
      title={R1-Searcher: Incentivizing the Search Capability in LLMs via Reinforcement Learning}, 
      author={Huatong Song and Jinhao Jiang and Yingqian Min and Jie Chen and Zhipeng Chen and Wayne Xin Zhao and Lei Fang and Ji-Rong Wen},
      year={2025},
      eprint={2503.05592},
      archivePrefix={arXiv},
      primaryClass={cs.AI},
      url={https://arxiv.org/abs/2503.05592}, 
}

@misc{zhang2025evolvesearchiterativeselfevolvingsearch,
      title={EvolveSearch: An Iterative Self-Evolving Search Agent}, 
      author={Dingchu Zhang and Yida Zhao and Jialong Wu and Baixuan Li and Wenbiao Yin and Liwen Zhang and Yong Jiang and Yufeng Li and Kewei Tu and Pengjun Xie and Fei Huang},
      year={2025},
      eprint={2505.22501},
      archivePrefix={arXiv},
      primaryClass={cs.CL},
      url={https://arxiv.org/abs/2505.22501}, 
}

@book{fuller1969morality,
  title={The Morality of Law},
  author={Fuller, Lon L.},
  year={1969},
  edition={Revised},
  publisher={Yale University Press},
  address={New Haven, CT},
}

@inproceedings{wang2024lekube,
    author = {Changyue Wang and Weihang Su and Yiran Hu and Qingyao Ai and Yueyue Wu and Cheng Luo and Yiqun Liu and Min Zhang and Shaoping Ma},
    title = {{LeKUBE}: A Knowledge Update {BEnchmark} for Legal Domain},
    booktitle = {Proceedings of the 2024 Annual International ACM SIGIR Conference on Research and Development in Information Retrieval in the Asia Pacific Region (SIGIR-AP '24)},
    year = {2024},
    pages = {175--185},
    publisher = {ACM},
    address = {Tokyo, Japan},
    doi = {10.1145/3673791.3698407},
    url = {https://doi.org/10.1145/3673791.3698407},
}

@inproceedings{t-y-s-s-etal-2024-chronoslex,
    title = "{C}hronos{L}ex: Time-aware Incremental Training for Temporal Generalization of Legal Classification Tasks",
    author = "T.y.s.s, Santosh  and Vuong, Tuan-Quang  and Grabmair, Matthias",
    booktitle = "Proceedings of the 62nd Annual Meeting of the Association for Computational Linguistics (Volume 1: Long Papers)",
    month = aug,
    year = "2024",
    address = "Bangkok, Thailand",
    publisher = "Association for Computational Linguistics",
    url = "https://aclanthology.org/2024.acl-long.166/",
    doi = "10.18653/v1/2024.acl-long.166",
    pages = "3022--3039",
}

@inproceedings{t-y-s-s-vuong-2025-lextempus,
    title = "{L}ex{T}empus: Enhancing Temporal Generalizability of Legal Language Models Through Dynamic Mixture of Experts",
    author = "T.y.s.s, Santosh  and Vuong, Tuan-Quang",
    booktitle = "Proceedings of the 63rd Annual Meeting of the Association for Computational Linguistics (Volume 1: Long Papers)",
    month = jul,
    year = "2025",
    address = "Vienna, Austria",
    publisher = "Association for Computational Linguistics",
    url = "https://aclanthology.org/2025.acl-long.329/",
    doi = "10.18653/v1/2025.acl-long.329",
    pages = "6608--6624",
}

@misc{yu2025dapoopensourcellmreinforcement,
      title={DAPO: An Open-Source LLM Reinforcement Learning System at Scale},
      author={Qiying Yu and Zheng Zhang and Ruofei Zhu and Yufeng Yuan and Xiaochen Zuo and Yu Yue and Weinan Dai and Tiantian Fan and Gaohong Liu and Lingjun Liu and Xin Liu and Haibin Lin and Zhiqi Lin and Bole Ma and Guangming Sheng and Yuxuan Tong and Chi Zhang and Mofan Zhang and Wang Zhang and Hang Zhu and Jinhua Zhu and Jiaze Chen and Jiangjie Chen and Chengyi Wang and Hongli Yu and Yuxuan Song and Xiangpeng Wei and Hao Zhou and Jingjing Liu and Wei-Ying Ma and Ya-Qin Zhang and Lin Yan and Mu Qiao and Yonghui Wu and Mingxuan Wang},
      year={2025},
      eprint={2503.14476},
      archivePrefix={arXiv},
      primaryClass={cs.LG},
      url={https://arxiv.org/abs/2503.14476},
}

@misc{shao2024deepseekmathpushinglimitsmathematical,
      title={DeepSeekMath: Pushing the Limits of Mathematical Reasoning in Open Language Models}, 
      author={Zhihong Shao and Peiyi Wang and Qihao Zhu and Runxin Xu and Junxiao Song and Xiao Bi and Haowei Zhang and Mingchuan Zhang and Y. K. Li and Y. Wu and Daya Guo},
      year={2024},
      eprint={2402.03300},
      archivePrefix={arXiv},
      primaryClass={cs.CL},
      url={https://arxiv.org/abs/2402.03300}, 
}

@misc{fan2025deepplannerscalingplanningcapability,
      title={DeepPlanner: Scaling Planning Capability for Deep Research Agents via Advantage Shaping}, 
      author={Wei Fan and Wenlin Yao and Zheng Li and Feng Yao and Xin Liu and Liang Qiu and Qingyu Yin and Yangqiu Song and Bing Yin},
      year={2025},
      eprint={2510.12979},
      archivePrefix={arXiv},
      primaryClass={cs.AI},
      url={https://arxiv.org/abs/2510.12979}, 
}

@article{huang2023lawyerllama,
  author       = {Quzhe Huang and Mingxu Tao and Zhenwei An and Chen Zhang and Cong Jiang and Zhibin Chen and Zirui Wu and Yansong Feng},
  title        = {Lawyer LLaMA Technical Report},
  journal      = {CoRR},
  volume       = {abs/2305.15062},
  year         = {2023},
  url          = {https://doi.org/10.48550/arXiv.2305.15062},
  doi          = {10.48550/ARXIV.2305.15062},
  eprinttype   = {arXiv},
  eprint       = {2305.15062},
}

@article{colombo2024saullm,
  author       = {Pierre Colombo and Telmo Pessoa Pires and Malik Boudiaf and Dominic Culver and Rui Melo and Caio Corro and Andr{\'{e}} F. T. Martins and Fabrizio Esposito and Vera L{\'{u}}cia Raposo and Sofia Morgado and Michael Desa},
  title        = {SaulLM-7B: {A} Pioneering Large Language Model for Law},
  journal      = {CoRR},
  volume       = {abs/2403.03883},
  year         = {2024},
  url          = {https://doi.org/10.48550/arXiv.2403.03883},
  doi          = {10.48550/ARXIV.2403.03883},
  eprinttype   = {arXiv},
  eprint       = {2403.03883},
}

@inproceedings{fei-etal-2025-internlm,
    title = "{I}ntern{LM}-Law: An Open-Sourced {C}hinese Legal Large Language Model",
    author = "Fei, Zhiwei  and
      Zhang, Songyang  and
      Shen, Xiaoyu  and
      Zhu, Dawei  and
      Wang, Xiao  and
      Ge, Jidong  and
      Ng, Vincent",
    editor = "Rambow, Owen  and
      Wanner, Leo  and
      Apidianaki, Marianna  and
      Al-Khalifa, Hend  and
      Eugenio, Barbara Di  and
      Schockaert, Steven",
    booktitle = "Proceedings of the 31st International Conference on Computational Linguistics",
    month = jan,
    year = "2025",
    address = "Abu Dhabi, UAE",
    publisher = "Association for Computational Linguistics",
    url = "https://aclanthology.org/2025.coling-main.629/",
    pages = "9376--9392"
}

@article{zhou2024lawgpt,
  author       = {Zhi Zhou and Jiang-Xin Shi and Peng-Xiao Song and Xiaowen Yang and Yi-Xuan Jin and Lan-Zhe Guo and Yufeng Li},
  title        = {LawGPT: {A} Chinese Legal Knowledge-Enhanced Large Language Model},
  journal      = {CoRR},
  volume       = {abs/2406.04614},
  year         = {2024},
  eprinttype   = {arXiv},
  eprint       = {2406.04614},
  url          = {https://arxiv.org/abs/2406.04614},
}

@article{cai2025unilawR1,
  author       = {Hua Cai and Shuang Zhao and Liang Zhang and Xuli Shen and Qing Xu and Weilin Shen and Zihao Wen and Tianke Ban},
  title        = {Unilaw-R1: {A} Large Language Model for Legal Reasoning with Reinforcement Learning and Iterative Inference},
  journal      = {CoRR},
  volume       = {abs/2510.10072},
  year         = {2025},
  eprinttype   = {arXiv},
  eprint       = {2510.10072},
  url          = {https://arxiv.org/abs/2510.10072},
}

@inproceedings{zhang2025syler,
  author       = {Kepu Zhang and Weijie Yu and Zhongxiang Sun and Jun Xu},
  title        = {SyLeR: A Framework for Explicit Syllogistic Legal Reasoning in Large Language Models},
  booktitle    = {Proceedings of the 34th ACM International Conference on Information and Knowledge Management},
  year         = {2025},
  publisher    = {ACM},
  doi          = {10.1145/3746252.3761120},
  url          = {https://doi.org/10.1145/3746252.3761120},
}

@inproceedings{li2024lexeval,
  author       = {Haitao Li and You Chen and Qingyao Ai and Yueyue Wu and Ruizhe Zhang and Yiqun Liu},
  title        = {LexEval: {A} Comprehensive Chinese Legal Benchmark for Evaluating Large Language Models},
  booktitle    = {Advances in Neural Information Processing Systems 38, NeurIPS 2024},
  year         = {2024},
  eprint       = {2409.20288},
  archivePrefix= {arXiv},
  primaryClass = {cs.CL},
  url          = {https://arxiv.org/abs/2409.20288},
}

@inproceedings{li2025webthinker,
  author       = {Xiaoxi Li and Jiajie Jin and Guanting Dong and Hongjin Qian and Yongkang Wu and Ji-Rong Wen and Yutao Zhu and Zhicheng Dou},
  title        = {WebThinker: Empowering Large Reasoning Models with Deep Research Capability},
  booktitle    = {Advances in Neural Information Processing Systems},
  year         = {2025},
  eprint       = {2504.21776},
  archivePrefix= {arXiv},
  primaryClass = {cs.AI},
  url          = {https://arxiv.org/abs/2504.21776},
}

@misc{chen2025researchlearningreasonsearch,
      title={ReSearch: Learning to Reason with Search for LLMs via Reinforcement Learning}, 
      author={Mingyang Chen and Linzhuang Sun and Tianpeng Li and Haoze Sun and Yijie Zhou and Chenzheng Zhu and Haofen Wang and Jeff Z. Pan and Wen Zhang and Huajun Chen and Fan Yang and Zenan Zhou and Weipeng Chen},
      year={2025},
      eprint={2503.19470},
      archivePrefix={arXiv},
      primaryClass={cs.AI},
      url={https://arxiv.org/abs/2503.19470}, 
}

@inproceedings{vu2024freshllms,
  title        = {{F}resh{LLM}s: Refreshing Large Language Models with Search Engine Augmentation},
  author       = {Tu Vu and Mohit Iyyer and Xuezhi Wang and Noah Constant and Jerry Wei and Jason Wei and Chris Tar and Yun-Hsuan Sung and Denny Zhou and Quoc Le and Thang Luong},
  booktitle    = {Findings of the Association for Computational Linguistics: ACL 2024},
  pages        = {13697--13720},
  publisher    = {Association for Computational Linguistics},
  year         = {2024},
  doi          = {10.18653/v1/2024.findings-acl.813},
  url          = {https://aclanthology.org/2024.findings-acl.813/},
}

@inproceedings{barale2025lextime,
  title        = {{L}ex{T}ime: A Benchmark for Temporal Ordering of Legal Events},
  author       = {Claire Barale and Leslie Barrett and Vikram Sunil Bajaj and Michael Rovatsos},
  booktitle    = {Findings of the Association for Computational Linguistics: EMNLP 2025},
  pages        = {5220--5236},
  publisher    = {Association for Computational Linguistics},
  year         = {2025},
  doi          = {10.18653/v1/2025.findings-emnlp.280},
  url          = {https://aclanthology.org/2025.findings-emnlp.280/},
}

@article{alzubi2025opendeepsearch,
  author       = {Salaheddin Alzubi and Creston Brooks and Purva Chiniya and Edoardo Contente and Chiara von Gerlach and Lucas Irwin and Yihan Jiang and Arda Kaz and Windsor Nguyen and Sewoong Oh and Himanshu Tyagi and Pramod Viswanath},
  title        = {Open Deep Search: Democratizing Search with Open-source Reasoning Agents},
  journal      = {CoRR},
  volume       = {abs/2503.20201},
  year         = {2025},
  eprinttype   = {arXiv},
  eprint       = {2503.20201},
  url          = {https://arxiv.org/abs/2503.20201},
}

@misc{verl,
  title={Volcano Engine Reinforcement Learning for LLMs},
  author={ByteDance Seed},
  year={2025},
  url={https://github.com/volcengine/verl},
}

@misc{zhou2026lras,
      title={LRAS: Advanced Legal Reasoning with Agentic Search},
      author={Yujin Zhou and Chuxue Cao and Jinluan Yang and Lijun Wu and Conghui He and Sirui Han and Yike Guo},
      year={2026},
      eprint={2601.07296},
      archivePrefix={arXiv},
      primaryClass={cs.AI},
      url={https://arxiv.org/abs/2601.07296},
}

@software{he2023hanfei,
  title={HanFei-1.0},
  author={He, Wanwei and Wen, Jiabao and Zhang, Lei and Cheng, Hao and Qin, Bowen and Li, Yunshui and Jiang, Feng and Chen, Junying and Wang, Benyou and Yang, Min},
  year={2023},
  publisher={GitHub},
  howpublished={\url{https://github.com/siat-nlp/HanFei}},
}

@misc{li2023sailerstructureawarepretrainedlanguage,
      title={SAILER: Structure-aware Pre-trained Language Model for Legal Case Retrieval}, 
      author={Haitao Li and Qingyao Ai and Jia Chen and Qian Dong and Yueyue Wu and Yiqun Liu and Chong Chen and Qi Tian},
      year={2023},
      eprint={2304.11370},
      archivePrefix={arXiv},
      primaryClass={cs.IR},
      url={https://arxiv.org/abs/2304.11370}, 
}

@inproceedings{ma2021lecard,
  title={LeCaRD: A Legal Case Retrieval Dataset for Chinese Law System},
  author={Ma, Yixiao and Shao, Yunqiu and Wu, Yueyue and Liu, Yiqun and Zhang, Ruizhe and Zhang, Min and Ma, Shaoping},
  booktitle={Proceedings of the 44th International ACM SIGIR Conference on Research and Development in Information Retrieval},
  pages={2342--2348},
  year={2021}
}

@misc{park2025lragelegalretrievalaugmented,
      title={LRAGE: Legal Retrieval Augmented Generation Evaluation Tool}, 
      author={Minhu Park and Hongseok Oh and Eunkyung Choi and Wonseok Hwang},
      year={2025},
      eprint={2504.01840},
      archivePrefix={arXiv},
      primaryClass={cs.CL},
      url={https://arxiv.org/abs/2504.01840}, 
}

@misc{wang2026lmarslegalmultiagentworkflow,
      title={L-MARS: Legal Multi-Agent Workflow with Orchestrated Reasoning and Agentic Search}, 
      author={Ziqi Wang and Boqin Yuan},
      year={2026},
      eprint={2509.00761},
      archivePrefix={arXiv},
      primaryClass={cs.AI},
      url={https://arxiv.org/abs/2509.00761}, 
}

\appendix
\crefalias{section}{appendix}

\section{Implementation Details}\label{app:hyperparameters}

We implement LegalSearch-R1 using Qwen2.5-7B-Instruct as the backbone model. For the auxiliary LLM used internally by the \texttt{browse\_webpage} and \texttt{rag\_retrieve} tools (for webpage content extraction and query analysis, respectively), we use Qwen3-30B-A3B~\cite{qwen2025qwen3technicalreport}. For agentic reinforcement learning, we adopt asynchronous multi-turn rollouts via VERL~\cite{verl} with FSDP parameter and optimizer offloading and Ulysses sequence parallelism (size 4). Training is conducted via full-parameter fine-tuning on a single node with 8 NVIDIA H100 GPUs. At evaluation time, we set the decoding temperature to 0 for deterministic greedy decoding with a single rollout per prompt. \cref{tab:training_config} summarizes the hyperparameters used for training.

\begin{table}[h]
\centering
\caption{Training configuration of LegalSearch-R1.}
\label{tab:training_config}
\begin{tabular}{l l}
\toprule
\textbf{Parameter} & \textbf{Value} \\
\midrule
Training batch size (global) & 64 \\
Concurrent rollouts ($G$) & 8 \\
Training steps & 112 \\
Learning rate & $1 \times 10^{-6}$ \\
EAS coefficient $\alpha$ & 0.1 \\
Clipping factor $\kappa$ & 2 \\
KL loss coefficient & 0 \\
Entropy coefficient & 0 \\
Clip ratio $\epsilon$ & 0.2 \\
Max prompt length & 4{,}096 \\
Max response length & 28{,}671 \\
Max context length & 32{,}767 \\
Max assistant turns & 15 \\
Rollout temperature & 1.0 \\
Top-$p$ & 1.0 \\
\bottomrule
\end{tabular}
\end{table}

\section{Tool Schema}\label{app:tool_schema}

\cref{figs:tool_schema} presents the tool schema configuration used by LegalSearch-R1 for external function calling. Each tool is defined with a name, a natural-language description that guides the agent's tool selection, and a structured parameter specification. The descriptions are domain-specific, explicitly scoping \texttt{web\_search} to legal information not covered by the local corpus, \texttt{rag\_retrieve} to Chinese statutory provisions across all temporal versions, and \texttt{browse\_webpage} to URLs obtained from prior search results. These schemas are loaded at initialization and provided to the agent as part of the system prompt.

\section{System Prompt}\label{app:system_prompt}

The system prompt establishes the agent's role as a professional Chinese legal research assistant and enforces strict tool-grounded reasoning. It explicitly declares that the agent's parametric legal knowledge may be incomplete or outdated, prohibiting the use of any information not obtained through tool calls. The prompt specifies the output format for each interaction round (think-plan in the first round, think-tool\_call in subsequent rounds, think-answer in the final round) and provides tool routing guidelines that direct statutory queries to \texttt{rag\_retrieve} and non-statutory legal questions to \texttt{web\_search}. The full prompt is shown in~\cref{figs:system_prompt}.

\section{Tool-Specific Prompts}\label{app:tool_prompts}

Beyond the agent's system prompt, two tools employ internal prompts to guide their processing. The \texttt{browse\_webpage} tool uses a legal reading agent with a domain-specific extraction prompt (\cref{figs:reading_prompt}) that enforces verbatim preservation of statutory text, complete extraction of judicial documents, and retention of temporal metadata such as effective dates and revision history. The \texttt{rag\_retrieve} tool uses an LLM-based query analyzer (\cref{figs:rag_prompt}) that extracts structured information from each query, including temporal references (converted to standardized date ranges), article/chapter identifiers (converted to Chinese numerals), and legal keywords, before the retrieval pipeline performs temporal filtering and hybrid search.

\section{Out-of-Domain Results}\label{app:ood_results}

To evaluate generalization beyond the training distribution, we test LegalSearch-R1 on six out-of-domain tasks drawn from professional legal examinations in DISC-LawEval~\cite{yue2023disclawllm}, spanning Certified Public Accountant (CPA), Patent Agent (PAE), National Judicial (NJE), Unified National Graduate Entrance (UNGEE), Legal Basic Knowledge (LBK), and Professional Fundamentals (PFE) examinations. These tasks are unseen during training and cover diverse areas of Chinese law beyond the criminal and civil domains of the in-domain benchmark. \cref{tabs:ood_evaluation_overall_performance} presents the full results.

\section{Task Examples}\label{app:task_examples}

We present one representative example per task to illustrate the scope, format, and difficulty of our benchmark. The in-domain examples (\cref{figs:id_examples}) cover seven tasks spanning factual knowledge recall, charge and sentencing prediction, statute recitation under temporal constraints, legal consultation, and case analysis. The out-of-domain examples (\cref{figs:ood_examples}) cover six professional examination tasks from diverse legal domains. All examples are shown with their ground-truth answers and have been translated from Chinese to English for accessibility.

\section{Annotation Guidelines for Legal Article Recitation}\label{app:lar_annotation}

Annotators are law students recruited from Chinese law schools and compensated at 15 USD per hour. The task requires identifying statutory provisions that undergo significant revisions across amendment versions of the Criminal Law, Criminal Procedure Law, and Civil Procedure Law. The full annotation instructions are shown in~\cref{figs:lar_annotation_guidelines}. The resulting corpus covers 13 amendment versions spanning 2000 to 2025, with each provision annotated with a temporal validity window ($t_{\text{from}}$, $t_{\text{to}}$) indicating the period during which the provision was legally effective.

\section{Case Study}\label{app:case_study}

We present a representative Legal Case Analysis~(LCA) example to illustrate how LegalSearch-R1 leverages the hybrid online-local retrieval architecture with temporal awareness. The agent first uses \texttt{web\_search} to gather general legal theory, then turns to \texttt{rag\_retrieve} for precise statutory provisions when web results prove insufficient for article-level reasoning.

\begin{table*}[t]
\renewcommand{\arraystretch}{1}
\small
\setlength{\tabcolsep}{4pt}
\begin{center}
\begin{tabular}{
    m{2.7cm} 
    m{0.5cm}<{\centering} 
    m{0.2cm}<{\centering} 
    m{0.7cm}<{\centering} 
    |m{0.7cm}<{\centering}m{0.7cm}<{\centering}
    m{0.7cm}<{\centering}m{1cm}<{\centering}m{0.7cm}<{\centering}m{0.7cm}<{\centering}
    |m{0.9cm}<{\centering} 
}
\toprule
\multirow{2}{*}{\textbf{Method}} &
\multicolumn{3}{c|}{\textbf{Setting}}&
\multirow{2}{*}{\textbf{CPA}} & \multirow{2}{*}{\textbf{PAE}} & \multirow{2}{*}{\textbf{NJE}} & \multirow{2}{*}{\textbf{UNGEE}} &  \multirow{2}{*}{\textbf{LBK}} & \multirow{2}{*}{\textbf{PFE}} & \multirow{2}{*}{\textbf{Avg.}} \\
&
\multirow{1}{*}{\textbf{\scriptsize Think}} &
\multirow{1}{*}{\textbf{\scriptsize Web}} &
\multirow{1}{*}{\textbf{\scriptsize RAG}} &&&&&&\\
\midrule
\multicolumn{11}{l}{\textbf{\textit{$\bullet$ Legal LLMs}}} \\
DISC-LawLLM       & \color{red!70!black}\ding{55} & \color{red!70!black}\ding{55} & \color{red!70!black}\ding{55} & 59.38 & 1.56 & 3.91 & 11.72 & 39.84 & 41.41 & 19.01  \\
DISC-LawLLM       & \color{green!70!black}\ding{52} & \color{red!70!black}\ding{55} & \color{red!70!black}\ding{55} & 39.06 & 10.94 &7.03  &35.94  & 51.56 & 64.06 & 34.77  \\
LegalDelta        & \color{red!70!black}\ding{55} & \color{red!70!black}\ding{55} & \color{red!70!black}\ding{55} & 56.25 & 40.62 & 34.38 & 53.12 & 45.31 & 50.78 &   46.74 \\
LegalDelta        & \color{green!70!black}\ding{52} & \color{red!70!black}\ding{55} & \color{red!70!black}\ding{55} & 59.38 & 36.72 & 34.38 &57.81  & 52.34 & 55.47 &  49.35 \\
\midrule

\multicolumn{11}{l}{$\spadesuit$ \textbf{\textit{Vanilla LLMs}}} \\
GPT-4o-mini        & \color{red!70!black}\ding{55} & \color{red!70!black}\ding{55} & \color{red!70!black}\ding{55} & 28.91  &  22.66  &  23.44  &  35.94  &  41.41  &  44.53  &  32.81  \\
\rowcolor{pink!30}
Qwen-2.5-7b        & \color{red!70!black}\ding{55} & \color{red!70!black}\ding{55} & \color{red!70!black}\ding{55} & 58.59  &  26.56  &  28.12  &  53.12  &  65.62  &  65.62  &  49.61 \\
Qwen-3-8b        & \color{red!70!black}\ding{55} & \color{red!70!black}\ding{55} & \color{red!70!black}\ding{55}  & 58.59  &  32.03  &  39.06  &  65.62  &  75.00  &  79.69  &  58.33\\
Qwen-3-14b        & \color{red!70!black}\ding{55} & \color{red!70!black}\ding{55} & \color{red!70!black}\ding{55}  & 71.09  &  42.97  &  41.41  &  61.72  &  76.56  &  84.38  &  63.02 \\
Qwen-3-30b-A3b     & \color{red!70!black}\ding{55} & \color{red!70!black}\ding{55} & \color{red!70!black}\ding{55} & 76.56  &  43.75  &  40.62  &  64.84  &  72.66  &  84.38  &  63.80 \\
\midrule

\multicolumn{11}{l}{$\clubsuit$ \textbf{\textit{LRMs}}} \\
o4-mini & \color{green!70!black}\ding{52} & \color{red!70!black}\ding{55} & \color{red!70!black}\ding{55} & 48.44  &  36.72  &  28.91  &  51.56  &  63.28  &  66.41  &  49.22 \\
QwQ-32b     & \color{red!70!black}\ding{55} & \color{red!70!black}\ding{55} & \color{red!70!black}\ding{55} & 75.00  &  33.59  &  38.28  &  53.91  &  60.94  &  81.25  &  57.16 \\
Qwen-3-30b-A3b     & \color{red!70!black}\ding{55} & \color{red!70!black}\ding{55} & \color{red!70!black}\ding{55} & 75.78  &  44.53  &  42.19  &  71.09  &  78.12  &  85.94  &  66.28 \\
DeepSeek-V3.2 & \color{green!70!black}\ding{52} & \color{red!70!black}\ding{55} & \color{red!70!black}\ding{55} & 79.69  &  42.19  &  42.97  &  65.62  &  80.47  &  87.50  &  66.41 \\
\midrule
\multicolumn{11}{l}{$\diamond$ \textbf{\textit{Deep Research}}} \\
Search-r1-base      & \color{green!70!black}\ding{52} & \color{red!70!black}\ding{55} & \color{green!70!black}\ding{52}  & 57.03  &  28.12  &  24.22  &  60.16  &  65.62  &  78.12  &  52.21  \\
Search-r1-instruct & \color{green!70!black}\ding{52} & \color{red!70!black}\ding{55} & \color{green!70!black}\ding{52} & 50.00  &  25.00  &  28.91  &  51.56  &  61.72  &  69.53  &  47.79 \\
DeepResearcher    & \color{green!70!black}\ding{52} & \color{green!70!black}\ding{52} & \color{red!70!black}\ding{55} & 52.34 & 23.43 & 27.34 & 54.69 & 60.94 & 71.88 & 48.43 \\
DeepPlanner    & \color{green!70!black}\ding{52} & \color{green!70!black}\ding{52} & \color{red!70!black}\ding{55} & 63.28  &  27.34  &  25.78  &  59.38  &  67.19  &  80.47  &  53.91 \\
\midrule
\rowcolor{cyan!15}
\textbf{LegalSearch-R1}    & \color{green!70!black}\ding{52} & \color{green!70!black}\ding{52} & \color{green!70!black}\ding{52}   & 70.31  &  42.19  &  41.41  &  70.31  &  74.22  &  83.59  &  63.67  \\
\bottomrule
\end{tabular}
\end{center}
\caption{Out-of-domain evaluation results on six legal examination tasks.}
\label{tabs:ood_evaluation_overall_performance}
\end{table*}

\begin{figure*}[t]
\begin{tcolorbox}[
  colback = cBlue_1!5,
  colframe = cBlue_6,
  coltitle = white,
  fonttitle = \bfseries\small,
  fontupper = \scriptsize,
  title = {§ System Prompt for LegalSearch-R1}
]
\textbf{Background.} Today's date is \{current\_date\}. You are a professional Chinese legal research assistant, but your legal knowledge may be incomplete or outdated. To ensure accuracy, you must use the available tools to retrieve authentic legal provisions, judicial interpretations, or authoritative information.\\[3pt]

\textbf{Task.} The user will pose questions related to Chinese law. You are prohibited from using any information not obtained through search tools for reasoning and answering. You must first formulate a systematic search plan, then call tools to retrieve authentic legal provisions, and finally provide a rigorous, traceable legal analysis based on the retrieved results.\\[3pt]

\textbf{Mandatory Requirements.}\\
1. In the first round, you must output your thinking process and a step-by-step research plan within \texttt{<think>} and \texttt{<plan>} tags.\\
2. Every round must begin with \texttt{<think>}, with no exceptions.\\
3. The final round must output your thinking process and final answer within \texttt{<think>} and \texttt{<answer>} tags.\\[3pt]

\textbf{Tool Routing Rules.}\\
1. If the question involves specific legal provisions (containing terms like ``Article X'', ``Criminal Law'', ``Civil Code''), use \texttt{rag\_retrieve}.\\
2. If the question involves legal theory, rule-of-law principles, institutional development, case analysis, or judicial practice, use \texttt{web\_search}.
\end{tcolorbox}
\caption{System prompt used during training and evaluation.}
\label{figs:system_prompt}
\end{figure*}

\begin{figure*}[t]
\begin{tcolorbox}[
  colback = cOrange!5,
  colframe = cOrange!80!black,
  coltitle = white,
  fonttitle = \bfseries\small,
  fontupper = \scriptsize,
  title = {§ Reading Agent Extraction Prompt (Browse Webpage)}
]
You are a professional and reliable Chinese legal research assistant. You will be given the user's main question, a current sub-question, context gathered so far, and one page of a webpage with its page index.\\[3pt]

Your task is to carefully read this page and extract all information that is \textbf{new relative to the existing context} and helpful for answering the main question or sub-question.\\[3pt]

\textbf{Special Requirements (Chinese Legal Research):}\\
1. When the page contains \textbf{original Chinese legal provisions} (from the Civil Code, Criminal Law, administrative regulations, judicial interpretations, etc.), preserve the text \textbf{verbatim} without paraphrasing.\\
2. When the page contains \textbf{judicial documents, penalty decisions, case details, or legal reasoning for applying specific articles}, extract them as completely as possible.\\
3. When the page contains \textbf{effective dates, version identifiers, or revision history}, these must be retained.\\
4. Content unrelated to the Chinese legal system may be skipped, but anything with potential relevance should be recorded.\\
5. \textbf{Scholarly interpretations, academic viewpoints, controversies, and expert opinions} should all be extracted.\\[3pt]

Output \texttt{<extracted\_info>}, \texttt{<page\_down>} (yes/no), and \texttt{<short\_summary>}.
\end{tcolorbox}
\caption{Domain-specific extraction prompt used by the reading agent in the \texttt{browse\_webpage} tool.}
\label{figs:reading_prompt}
\end{figure*}

\begin{figure*}[t]
\begin{tcolorbox}[
  colback = cOrange!5,
  colframe = cOrange!80!black,
  coltitle = white,
  fonttitle = \bfseries\small,
  fontupper = \scriptsize,
  title = {§ RAG Query Analysis Prompt (rag\_retrieve)}
]
Extract the following structured information from the given Chinese query and output in standard JSON format. Do not output any explanation or extra text.\\[3pt]

\textbf{1. Temporal Information} (\texttt{time\_info})\\
Format: \texttt{YYYY-MM-DD}. Expand partial dates to start/end ranges (e.g., ``2024'' $\to$ [\texttt{"2024-01-01"}, \texttt{"2024-12-31"}]). Extract all time periods mentioned. Return empty list if none.\\[3pt]

\textbf{2. Chapter/Article References} (\texttt{chapter\_info})\\
Extract references such as ``Article X'' or ``Chapter X'' in full Chinese numeral form. Convert Arabic numerals to Chinese numerals (e.g., ``Criminal Law Art. 3'' $\to$ ``Article Three''). Return empty list if none.\\[3pt]

\textbf{3. Keywords} (\texttt{keywords})\\
Extract primary legal concepts and terms. Must be semantically complete words that appear in the query. Include sub-terms of compound words (e.g., ``intentional homicide'' should also include ``intentional'').\\[3pt]

\textbf{Output Format:} \texttt{\{"time\_info": [], "chapter\_info": [], "keywords": []\}}
\end{tcolorbox}
\caption{Structured query analysis prompt used by the RAG retrieval pipeline to extract temporal context, article references, and keywords before retrieval.}
\label{figs:rag_prompt}
\end{figure*}

\begin{figure*}[t]
\begin{tcolorbox}[
  colback = cBlue_1!5,
  colframe = cBlue_6,
  coltitle = white,
  fonttitle = \bfseries\small,
  fontupper = \scriptsize,
  title = {§ Tool Schema Configuration (YAML)}
]
\begin{verbatim}
tools:
  - name: "web_search"
    description: "Search the internet for legal information related to the
      user's question and legal provisions not covered by rag_retrieve."
    parameters:
      query:
        type: array
        description: "List of query strings, e.g. Article 3 of the 2014
          Administrative Penalty Law."
        items: {type: string}

  - name: "browse_webpage"
    description: "Browse specific webpages and extract legal-related content."
    parameters:
      url_list:
        type: array
        description: "List of URLs to browse. URLs must come from prior
          web_search results."
        items: {type: string}

  - name: "rag_retrieve"
    description: "Retrieve Chinese legal provisions from the local corpus,
      covering all temporal versions of 16 statutes including Criminal Law,
      Civil Code, Criminal/Civil Procedure Law, and others."
    parameters:
      query:
        type: array
        description: "List of query strings, e.g. 2015 intentional homicide."
        items: {type: string}
\end{verbatim}
\end{tcolorbox}
\caption{Tool schema configuration for LegalSearch-R1.}
\label{figs:tool_schema}
\end{figure*}

\begin{figure*}[t]
\begin{tcolorbox}[
  colback = cBlue_1!5,
  colframe = cBlue_6,
  coltitle = white,
  fonttitle = \bfseries\small,
  fontupper = \scriptsize,
  title = {§ In-Domain Task Examples}
]

\textbf{KQA (Knowledge Question Answering)}\\
\textit{Q.} Which of the following statements about international factoring are correct? A: Factoring mainly includes four services\ldots\ B: The export factor and import factor are bound by a mutual factoring contract\ldots\ C: The importer and exporter are bound by a goods sale contract\ldots\ D: The exporter and export factor are bound by a contract established under the export factoring agreement.\\
\textit{Answer:} ABCD\\[4pt]

\textbf{CCP (Criminal Charge Prediction)}\\
\textit{Q.} Case facts: The defendant Zhang used forged identity documents of Wu and L\"u to fraudulently obtain 6 debit cards from six different banks.\\
\textit{Answer:} Obstructing credit card management\\[4pt]

\textbf{PTP (Prison Term Prediction)}\\
\textit{Q.} Case facts: The defendant Ke organized unauthorized burning of a mountain for land clearing. Due to wind, the fire spread beyond the intended area, burning 97.5 mu (6.5 hectares) of forested land with direct economic losses of 42{,}000 RMB. He voluntarily surrendered and compensated the victims. Relevant statute: Criminal Law Art.~115.\\
\textit{Answer:} 12 months\\[4pt]

\textbf{LAR (Legal Article Recitation)}\\
\textit{Q.} It is now March 2004. Please recite in full the text of Article~46 of the Criminal Procedure Law currently in effect.\\
\textit{Answer:} All case judgments shall emphasize evidence and investigation, and shall not readily credit confessions. A defendant may not be found guilty and sentenced based solely on confession without other evidence; a defendant may be found guilty and sentenced when evidence is sufficient and reliable even without confession.\\[4pt]

\textbf{LAP (Legal Article Prediction)}\\
\textit{Q.} Case facts: During the prohibited hunting season, defendant Li used bird nets to catch 26 quails, a species under state wildlife protection. All quails were subsequently released. Options: A: Criminal Law Art.~89 \; B: Art.~237 \; C: Art.~63 \; D: Art.~341.\\
\textit{Answer:} D\\[4pt]

\textbf{LCA (Legal Case Analysis)}\\
\textit{Q.} Wang and Li conspired to rob a pedestrian of 10{,}000 RMB under identical circumstances. The court sentenced Wang (son of a county official) to 3 years and Li to 5 years. Which principle did the court violate? A: Principle of legality \; B: Proportionality of crime and punishment \; C: Combined punishment and education \; D: Equality before criminal law.\\
\textit{Answer:} D\\[4pt]

\textbf{LCS (Legal Consultation)}\\
\textit{Q.} I joined a Shanghai company in April. The job posting stated business trips of 15 days to one month, and the contract specifies Shanghai as the primary workplace. How is severance calculated?\\
\textit{Answer:} An employee who voluntarily resigns generally receives no severance; however, if the employer violated the employee's rights, severance is owed. Under Labor Law Art.~47, severance is one month's salary per year of service\ldots
\end{tcolorbox}
\caption{Representative examples of the 7 in-domain tasks in our benchmark, covering factual knowledge, charge prediction, sentencing, statute recitation, article prediction, case analysis, and legal consultation.}
\label{figs:id_examples}
\end{figure*}

\begin{figure*}[t]
\begin{tcolorbox}[
  colback = cOrange!5,
  colframe = cOrange!80!black,
  coltitle = white,
  fonttitle = \bfseries\small,
  fontupper = \scriptsize,
  title = {§ Out-of-Domain Task Examples}
]

\textbf{CPA (Certified Public Accountant Examination)}\\
\textit{Q.} [Year 2015] A coal mining enterprise sold 50{,}000 tons of washed coal in April 2015 for 50 million RMB. The wash-to-raw coal conversion rate is 80\% and the resource tax rate is 10\%. How much resource tax is owed (in ten-thousands of RMB)? A: 400 \; B: 404 \; C: 505 \; D: 625.\\
\textit{Answer:} A\\[4pt]

\textbf{PAE (Patent Agent Examination)}\\
\textit{Q.} [Year 2021] In which of the following cases is the patent applicant entry non-compliant? A: A company abbreviated its name \; B: A university listed its research office \; C: A foreign professor appended ``Professor'' to his name \; D: An individual used a pen name.\\
\textit{Answer:} ABCD\\[4pt]

\textbf{NJE (National Judicial Examination)}\\
\textit{Q.} [Year 2019] Xiao Zhang purchased Ferrero chocolates online for 1{,}000 RMB, advertised as Italian imports, but the packaging listed a Beijing manufacturer. Which statements are correct? A: Zhang may return unopened items within 7 days \; B: Return shipping is borne by the seller \; C: Zhang may claim 3{,}000 RMB in damages \; D: Zhang may claim 10{,}000 RMB in damages.\\
\textit{Answer:} AC\\[4pt]

\textbf{UNGEE (Unified National Graduate Entrance Examination)}\\
\textit{Q.} [Year 2017] Couple A and B have one son C. C married D and had a daughter E. C died in 2008. D remarried G, and together they cared for A and B. A died in May 2015. Who are the first-order heirs to A's estate? A: B \; B: D \; C: E \; D: G.\\
\textit{Answer:} ABC\\[4pt]

\textbf{LBK (Legal Basic Knowledge)}\\
\textit{Q.} What is the warranty against rights defects in a sales contract? A: The seller's delivered goods shall meet quality descriptions \; B: The buyer may notify quantity discrepancies within two years \; C: The buyer may notify quality issues within warranty \; D: The seller warrants that no third party will make claims against the buyer.\\
\textit{Answer:} D\\[4pt]

\textbf{PFE (Professional Fundamentals Examination)}\\
\textit{Q.} [Year 2017] Which of the following is NOT a civil servant? A: A county party committee member \; B: A municipal administration officer \; C: A state-owned enterprise worker \; D: A democratic party functionary.\\
\textit{Answer:} C
\end{tcolorbox}
\caption{Representative examples of the 6 out-of-domain tasks drawn from professional legal examinations, each with an explicit temporal context embedded in the question.}
\label{figs:ood_examples}
\end{figure*}

\begin{figure*}[t]
\begin{tcolorbox}[
  colback = cBlue_1!5,
  colframe = cBlue_6,
  coltitle = white,
  fonttitle = \bfseries\small,
  fontupper = \small,
  title = {Annotation Guidelines for Legal Article Recitation}
]
\textbf{Objective:} Identify statutory provisions that undergo significant revisions across amendment versions of the Criminal Law, Criminal Procedure Law, and Civil Procedure Law.\\[4pt]

\textbf{Selection Criteria} (satisfy at least one):\\
\begin{enumerate}
\item The revised text differs from the original by \textbf{more than 20\% in character count}.
\item The revision introduces a \textbf{clear semantic change} that alters the legal meaning or applicability of the provision (\eg addition or removal of qualifying conditions, changes in sentencing ranges, reversal of legal presumptions).
\end{enumerate}
\vspace{2pt}

\textbf{Data Format:}\\[2pt]
\begin{tabular}{@{}p{0.47\textwidth} p{0.47\textwidth}@{}}
\textit{Pre-revision entry:} & \textit{Post-revision entry:} \\
\quad $\bullet$ Article number & \quad $\bullet$ Revised article number (if renumbered) \\
\quad $\bullet$ Amendment date (effective date) & \quad $\bullet$ Amendment date (effective date) \\
\quad $\bullet$ Full article text & \quad $\bullet$ Full revised article text \\
\end{tabular}
\vspace{4pt}

\end{tcolorbox}
\caption{Annotation guidelines provided to law student annotators for the Legal Article Recitation~(LAR) task.}
\label{figs:lar_annotation_guidelines}
\end{figure*}

\begin{figure*}[t]
\begin{tcolorbox}[
  colback = cBlue_1!5,
  colframe = cBlue_6,
  coltitle = white,
  fonttitle = \bfseries\small,
  fontupper = \scriptsize,
  fontlower = \scriptsize,
  title = {§ Legal Case Analysis: Inheritance Dispute with Conflicting Wills (LCA Task, Answer: C)}
]
\textbf{Question:} A couple made three conflicting wills: a holographic will (2001) dividing three rooms among their children, a \textit{notarized will} (2002) reassigning one room, and an \textit{audio-recorded will} (2004) reassigning another room. Both spouses passed away in 2004. Which will should the court follow? Options: A: Audio-recorded will \; B: Holographic will \; C: Notarized will \; D: Statutory succession\\[4pt]
\hrule\vspace{4pt}

\texttt{<think>} The question examines the legal priority among different will forms. The events occurred in 2001--2004, so I need to determine which law was in effect at that time. I should first search for relevant legal theory on will priority rules. \texttt{</think>}\\
\texttt{<plan>} 1. Search for legal theory on the priority between notarized and audio-recorded wills, paying attention to temporal changes in the law. 2. Retrieve precise statutory text from the RAG corpus with the correct temporal context (2004). 3. Apply the temporally appropriate statute to determine the answer. \texttt{</plan>}\\[4pt]

\texttt{<think>} Following the plan, I will first search for general legal theory on will form priority. \texttt{</think>}\\
\texttt{<tool\_call>} \texttt{\{"name": "web\_search", "arguments": \{"query": ["priority between audio-recorded will and notarized will", "applicable order of will forms"]\}\}} \texttt{</tool\_call>}\\[2pt]
\texttt{<tool\_response>} \textit{[1] ``Civil Code's Impact on Will Priority Rules -- LegalDaily.cn'' --- After the Civil Code took effect on January 1, 2021, notarized wills no longer enjoy automatic priority over other will forms. The new Art.~1142 establishes that the last will in time prevails... [2] ``How to Determine Will Validity: Notarized vs. Other Forms -- zhihu.com'' --- Before the Civil Code, notarized wills had absolute priority under Art.~20 of the Inheritance Law (1985). Other forms of wills could not revoke a notarized will... [3] ``Analysis of Inheritance Disputes Involving Multiple Wills -- sohu.com'' --- Courts must first determine the applicable law based on when the testamentary acts occurred...} \texttt{</tool\_response>}\\[4pt]

\texttt{<think>} Web search reveals that the legal priority of notarized wills changed with the Civil Code (effective 2021). Since the case events occurred in 2001--2004, the old Inheritance Law should apply. However, web results lack precise statutory provisions. I need to retrieve the exact legal text via RAG with temporal context. \texttt{</think>}\\
\texttt{<tool\_call>} \texttt{\{"name": "rag\_retrieve", "arguments": \{"query": ["2004 notarized will", "2004 audio-recorded will"]\}\}} \texttt{</tool\_call>}\\[2pt]
\texttt{<tool\_response>} \textit{[RAG returns: Civil Code Art.~1142 (eff. 2021.01.01): ``The testator may revoke or modify a will. When multiple wills conflict, the last will prevails.'' Also Art.~1139: ``A notarized will shall be processed by a notarization institution.'' And the old Inheritance Law Art.~20 (eff. 1985--2020): ``Holographic, dictated, audio-recorded, and oral wills may not revoke or alter a notarized will.'']} \texttt{</tool\_response>}\\[4pt]

\texttt{<think>} The RAG results confirm the temporal distinction. Under the old Inheritance Law (applicable to events in 2001--2004), Art.~20 explicitly states that audio-recorded wills cannot revoke or alter a notarized will. The 2004 audio-recorded will therefore cannot override the 2002 notarized will. The court should divide the estate according to the notarized will. \texttt{</think>}\\[2pt]
\texttt{<answer>} \colorbox{green!15}{\textbf{C}} \texttt{</answer>} \quad {\color{green!50!black}\ding{52} Correct}
\end{tcolorbox}
\caption{A case study illustrating how LegalSearch-R1 combines \texttt{web\_search} for general legal analysis with \texttt{rag\_retrieve} for precise statute retrieval, and applies temporal reasoning to identify the correct legal regime (pre-Civil Code Inheritance Law) based on when the events occurred.}
\label{figs:case_study}
\end{figure*}

\end{document}